\documentclass[review]{elsarticle}

\usepackage{lineno,hyperref}
\usepackage{amsmath,amsfonts,amssymb}
\usepackage{booktabs}
\usepackage{color}
\modulolinenumbers[5]
\usepackage{pifont}
\newcommand{\xmark}{\ding{55}}

\journal{and accepted by ISPRS Journal of Photogrammetry and Remote Sensing}

\begin{document}

\begin{frontmatter}

\title{MiniNet: An extremely lightweight convolutional neural network for real-time unsupervised monocular depth estimation}


\author[mymainaddress,mysecondaryaddress,3]{Jun Liu}

\author[mymainaddress,mysecondaryaddress,3]{Qing Li}

\author[mymainaddress,mysecondaryaddress,3]{Rui Cao}

\author[mymainaddress,mysecondaryaddress,3]{Wenming Tang}

\author[mymainaddress,mysecondaryaddress,3,4]{Guoping Qiu\corref{mycorrespondingauthor}}
\cortext[mycorrespondingauthor]{Corresponding author}
\ead{guoping.qiu@nottingham.ac.uk}

\address[mymainaddress]{College of Electronics and Information Engineering, Shenzhen University, Shenzhen, China}
\address[mysecondaryaddress]{Guangdong Key Laboratory of Intelligent Information Processing, Shenzhen University, Shenzhen, China}
\address[3]{Shenzhen Institute of Artificial Intelligence and Robotics for Society, Shenzhen, China}
\address[4]{School of Computer Science, The University of Nottingham, UK}

\begin{abstract}
Predicting depth from a single image is an attractive research topic since it provides one more dimension of information to enable machines to better perceive the world. Recently, deep learning has emerged as an effective approach to monocular depth estimation. As obtaining labeled data is costly, there is a recent trend to move from supervised learning to unsupervised learning to obtain monocular depth. However, most unsupervised learning methods capable of achieving high depth prediction accuracy will require a deep network architecture which will be too heavy and complex to run on embedded devices with limited storage and memory spaces. To address this issue, we propose a new powerful network with a recurrent module to achieve the capability of a deep network while at the same time maintaining an extremely lightweight size for real-time high performance unsupervised monocular depth prediction from video sequences. Besides, a novel efficient upsample block is proposed to fuse the features from the associated encoder layer and recover the spatial size of features with the small number of model parameters. We validate the effectiveness of our approach via extensive experiments on the KITTI dataset. Our new model can run at a speed of about 110 frames per second (fps) on a single GPU, 37 fps on a single CPU, and 2 fps on a Raspberry Pi 3. Moreover, it achieves higher depth accuracy with nearly 33 times fewer model parameters than state-of-the-art models. To the best of our knowledge, this work is the first extremely lightweight neural network trained on monocular video sequences for real-time unsupervised monocular depth estimation, which opens up the possibility of implementing deep learning-based real-time unsupervised monocular depth prediction on low-cost embedded devices.
	
\end{abstract}


\end{frontmatter}


\section{Introduction}
Estimating depth of surrounding scenes plays a crucial role in enabling machines to better perceive the world, which is key to robots, unmanned aerial vehicles (UAV) and wearable devices. It is also an important topic in photogrammetry and remote sensing applications such as visual odometry~\cite{RN718}, image localization~\cite{RN713}, and height estimation~\cite{RN720}. Currently, LiDAR, structured light depth sensors and time-of-flight sensors are employed to capture the depth information~\cite{RN315, RN721}. These active depth sensors are often heavy, expensive and power-consuming. Meanwhile, they suffer from noise and artifacts especially from the interferences of reflective or transparent surfaces. Besides, depth information can also be obtained from depth-from-defocus~\cite{RN726,RN464}, multi-view stereo (MVS)~\cite{RN719,RN722}, and structure from motion (SfM)~\cite{RN718} approaches. However, these approaches either are time-consuming or suffer from low depth accuracy. Thus, depth estimation using a single image from a RGB camera is an attractive alternative to the aforementioned depth estimation approaches, due to its compact, cheap and low-power properties.

\begin{figure*}[h]
	\centering
	\makebox[\textwidth][c]{\includegraphics[scale=0.83]{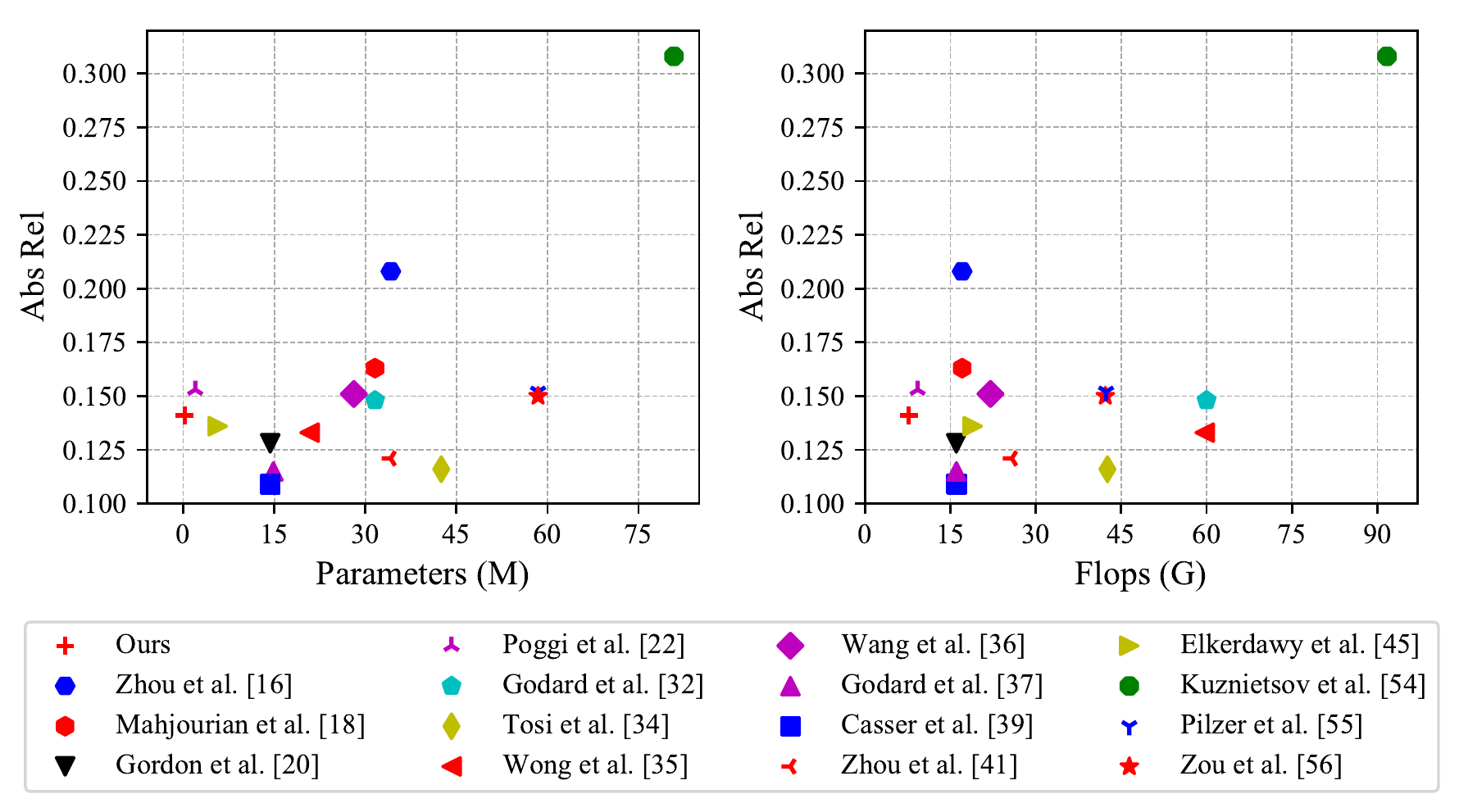}}
	\caption{Mean absolute relative error (Abs Rel) (see Section~\ref{section_DE} for more detail explanation) versus the model parameters (left view) and floating point operations per second (flops) (right view) on $ 640 \times 192 $ inputs for different unsupervised monocular estimation methods. M and G indicate $ \times 10^{6} $ and $ \times 10^{9} $, respectively. Best viewed in color.}
	\label{fig:0}
\end{figure*}

In the last decade, inspired by the success of deep learning in high-level vision tasks~\cite{RN192,RN268}, much research efforts have been directed towards supervised learning-based monocular depth estimation.  It casts the monocular depth estimation as a pixel-level regression problem and achieves impressive performance~\cite{RN284,RN330,RN413}. However, supervised learning methods rely on large labeled RGB-D datasets, which are expensive and burdensome. To circumvent the need of large labeled datasets, unsupervised approaches to monocular depth estimation have recently emerged in the literatures. These methods mimic the human binocular or monocular vision capabilities. The ground-truth depth-based loss is therein replaced by the image reconstruction loss~\cite{RN288,RN291,RN292}. Unlike binocular vision techniques have calibrated camera pose, monocular vision techniques have unknown and inconstant pose information between adjacent video frames. Thus, these techniques need additional convolutional neural networks (CNNs) for pose estimation. Since monocular video sequences are much more accessible than stereo image pairs, the unsupervised method trained on monocular video sequences is adopted in this paper to broaden the application range of our approach. However, most unsupervised learning methods require deep and complicated neural networks to achieve high estimation accuracy (e.g.~\cite{RN582,RN293,RN631}) as shown in Fig.~\ref{fig:0}, which illustrates the depth estimation error versus model parameters and computational speed requirements for various methods in the literature. It is seen that many existing methods have very large number of parameters and demand very high computational capabilities, which make them infeasible to implement on embedded devices with limited storage and memory spaces.

To address the issues, in this paper, we propose a compact and very effective unsupervised learning-based neural network for monocular depth estimation (named MiniNet), which effectively reduces parameters and floating point operations per second (flops) and meanwhile remains relatively high depth prediction accuracy. Our proposed model is composed of a DepthNet and two shared-weight PoseNets as shown in Fig.~\ref{fig:1}. Since only the DepthNet is used in the inference stage, we focus on the design of the DepthNet. Here, a recurrent module is proposed to construct the encoder of our DepthNet, which requires an identical input and output channel size for this recurrent module. The size of features is halved after each recurrently passing through this module. Thus, our encoder of DepthNet can achieve the effects of deep CNNs but with extremely lightweight parameters. Besides, a novel efficient upsample block is proposed to further reduce the parameters and flops of the DepthNet, which is mainly made up of depth-wise separable convolution~\cite{RN697} with a shortcut. It is adopted to upsample the feature maps and fuse the features from the corresponding encoder layer. Thanks to the lightweight encoder and new efficient decoder, the parameters of our DepthNet are about 9 times fewer than that of Poggi et al.~\cite{RN572}.

Our major contributions of this paper are summarized as follows:
\begin{enumerate}
\item We present an extremely lightweight deep learning-based model (named MiniNet) for unsupervised monocular depth estimation, where a recurrent module and a novel efficient upsample block are proposed. Our MiniNet can achieve real-time performance and meanwhile obtain very competitive depth prediction accuracy. To the best of our knowledge, our approach is the first extremely lightweight model trained on monocular video sequences for unsupervised depth estimation.
\item We propose a small version of our MiniNet, which can achieve real-time performance on both GPU and CPU cards. It can reach about 110 frames per second (fps) on a single GPU, 37 fps on a CPU only machine, and 2 fps on a Raspberry Pi 3. Moreover, the parameters of this model are about 33 times fewer than that of the one using eighth output resolution in Poggi et al.~\cite{RN572} and at the same time our model has better depth accuracy.
\item We have conducted extensive experiments on the KITTI dataset~\cite{RN315} to demonstrate the effectiveness and efficiency of our proposed models. Furthermore, we also conduct the experiment on the Make3D dataset~\cite{RN325} without fine-tuning on it, which demonstrates the good generalization ability of our MiniNet.
\end{enumerate}




The rest of the paper is organized as follows. Section~\ref{section_RW} reviews the related works. Section~\ref{section_ME} introduces the architecture of our proposed MiniNet, especially the structure of the DepthNet. Section~\ref{section_EX} presents experimental results on the KITTI and Make3D datasets. Finally, Section~\ref{section_CO} draws the conclusions of our work. 

\section{Related work} \label{section_RW}
In this section, we review the relevant works that take a single RGB image as input and estimate the depth value of each pixel as output at test time. These works can be categorized into supervised and unsupervised depth estimation according to whether to use the ground-truth depth or not at training time. Besides, the works about lightweight networks for monocular depth estimation tasks are also summarized in the last part of this section.

\subsection{Supervised depth estimation} 
In the earlier works, Saxena et al.~\cite{RN325} leveraged a Markov Random Field (MRF) trained by supervised learning to infer the image depth. It suffers from the lack of thin structures and global context information due to its local nature. Liu et al.~\cite{RN294} proposed a simpler MRF model to infer the depth map by using the predicted semantic labels from its first phase, where the Make3D dataset~\cite{RN325} with hand-annotated semantic class labels was employed for training and testing. Ladicky~\cite{RN331} jointly predicted the depth map and semantic segmentation labels by a pixel-wise classifier trained by the ground-truth depth and semantic information, while Liu et al.~\cite{RN323} recast the depth prediction as a discrete-continuous graphical model optimization problem by using image superpixels.

In a seminal work, Eigen et al.~\cite{RN268} are the first to adopt a CNN architecture for monocular depth estimation task, which consists of a coarse-scale network performing a global prediction and a fine-scale network refining predictions locally. Then, they~\cite{RN321} extended this approach to handle three correlative tasks, i.e. depth, surface normal, and semantic label predictions. Laina et al.~\cite{RN284} proposed a deeper residual network (ResNet50-type~\cite{RN192} encoder and Up-projection decoder) with a novel reverse Huber loss for monocular depth estimation. Then, it was found that probabilistic graphical models based CNNs are capable of boosting the performance of monocular depth estimation. Li et al.~\cite{RN337} proposed a hierarchical conditional random field (CRF) as a post-processing operation to refine the output depth map, while Liu et al.~\cite{RN283} integrated continuous CRF in a unified deep CNN framework for depth prediction. Xu et al.~\cite{RN329} introduced the CNN-implemented continuous CRF for aggregating the multi-scale feature maps. Apart from these end-to-end depth regression approaches, Fu et al.~\cite{RN413} formulated the depth estimation as an ordinal regression problem with their spacing-increasing discretization (SID) strategy to discretize depth and further improved the performance of depth estimation by a large margin.

Although supervised depth estimation is able to achieve high depth prediction accuracy, it requires large scale ground-truth depth labels, which come from expensive laser scanner or time-of-flight sensors. To get rid of this issue, unsupervised depth estimation is adopted in this paper. 

\subsection{Unsupervised depth estimation}
Another trend to monocular depth estimation is unsupervised learning, where the image reconstruction from stereo image pairs or monocular video sequences is treated as the supervisory signal and the depth map is the intermediate product. With synchronized stereo images in the training stage, Xie et al.~\cite{RN417} obtained discretized depth from a soft disparity map by minimizing the reconstruction error between the right view and generated right one from the left view. Garg et al.~\cite{RN288} extended this approach to output continuous depth value, but their image formation model was not fully differentiable and thus hard to optimize. Godard et al.~\cite{RN289} employed a bilinear sampler from the spatial transformer network (STN)~\cite{RN591} for full differentiable operation, and first introduced the left-right consistency of stereo images for training a depth estimation network. Tosi et al.~\cite{RN533} designed a new deep architecture for monocular depth estimation by synthesizing features from a different point of view as input to disparity refinement model and proposed a proxy ground truth annotation via traditional knowledge from stereo, i.e. Semi-Global Matching (SGM). Wong et al.~\cite{RN565} introduced a bilateral cyclic consistency constraint to enforce consistency between the left and right disparities and removed stereo dis-occlusions. Moreover, they proposed a model-driven adaptive weighting scheme to better balance data fidelity and regularization, which is also adopted in our loss function.    

On the other hand, monocular video sequences are used in training stage. In the earlier works, Zhou et al.~\cite{RN291} put forward a monocular depth estimation network with a multi-view camera ego-motion (camera pose) network using a monocular video. Mahjourian et al.~\cite{RN582} proposed a 3D point cloud alignment loss to further constrain the geometry consistency between consecutive video sequences. Wang et al.~\cite{RN350} introduced a normalization trick to address the scale sensitive issue and a differentiable direct visual odometry (DVO) to improve the performance of depth estimation. Yin et al.~\cite{RN293} proposed an adaptive geometric consistency loss to resolve occlusion and texture ambiguities problem by jointly unsupervised learning depth, optical flow and camera ego-motion from monocular videos. Godard et al.~\cite{RN435} designed the pixel-level minimum reprojection loss and auto-masking loss to handle the occlusions and stationary pixels. Bozorgtabar et al.~\cite{RN680} aligned the monocular depth estimation trained on unlabeled monocular videos with the deep features from synthetic images to resolve the scale ambiguity. These deep features were coupled with scene depth information. Casser et al.~\cite{RN668} better handled the moving objects by using pre-computed instance segmentation masks and imposing object size constraints. Gordon et al.~\cite{RN631} proposed to learn the camera intrinsic parameters in the unsupervised monocular depth estimation for the first time, addressed the occlusions differentiably, and introduced a randomized layer normalization for resolving objection motion issue. Ranjan et al.~\cite{RN575} jointly trained depth, camera ego-motion, optical flow and segmentation of static and moving regions, and introduced competitive collaboration to reinforce each other. Zhou et al.~\cite{RN682} proposed a dual network consisting of LR-Net, HR-Net and SA-Attention module to deal with high-resolution image efficiently.

In general, stereo image pairs are not as widely available as monocular video sequences, which are more easily collected. To broaden the applicability of our method, monocular video sequences are adopted to train our unsupervised depth estimation neural network.  

\subsection{Lightweight CNN for depth estimation}
Thanks to the development of the aforementioned works, the accuracy of depth prediction from monocular videos is comparable with that of the methods trained on stereo image pairs. However, they normally require a more sophisticated and deeper network architecture. To satisfy the requirement of practical application with limited storage space and computation resources, we need to further reduce the parameters and flops of the depth CNN, and meanwhile constrain the decrease of depth accuracy within a reasonable range. There are a few relevant works dedicated to realizing lightweight real-time depth estimation. For the supervised depth estimation, Wofk et al.~\cite{RN709} proposed a supervised-learning FastDepth, which was composed of a MobileNet-V1~\cite{RN697} based encoder and a depth-wise decomposition-based decoder. Using network pruning with the input size of $ 224 \times 224 $, they obtained 27 fps on a Jetson TX2 CPU with 1.34 M parameters. Besides, Nekrasov et al.~\cite{RN708} achieved the real-time performance for supervised depth estimation and semantic segmentation on a GT1080Ti GPU card, via a lightweight RefineNet architecture built on top of the MobileNet-v2~\cite{RN650}. They obtained 6.45 G flops on the input size of $ 1200 \times 350 $ with the parameters of 2.99 M.

On the other hand, for the unsupervised depth estimation, Poggi et al.~\cite{RN572} proposed a pyramidal structure-based unsupervised monocular depth estimation network trained on stereo image pairs. They obtained the real-time performance both on a standard GPU card with 1.972 M parameters and on a CPU card using eighth resolution output. Besides, Elkerdawy et al.~\cite{RN645} introduced an end-to-end filter pruning method likewise trained on stereo image pairs. It learned a binary mask to prune the large trained model and yielded 5.700 M model parameters. In this paper, we propose a much more lightweight network (named MiniNet) with 0.217 M parameters for the depthNet, which is the minimum among the unsupervised learning methods as shown in Fig.~\ref{fig:0}. Under the input image size of $ 640 \times 192 $, our MiniNet is able to realize the real-time performance on a standard GPU card. Furthermore, the small version of MiniNet is proposed to achieve the real-time performance of about 110 fps on a single GPU card and 37 fps on a single CPU card, as well as about 2 fps on a Raspberry Pi 3. Moreover, it has higher depth prediction accuracy and approximately 33 times fewer parameters than the state-of-the-art real-time unsupervised model~\cite{RN572}. To the best of our knowledge, we are the first one to propose a lightweight unsupervised monocular depth estimation network trained on monocular video sequences, which is more suitable for the usual environment with real-time performance and small storage requirements.

\section{Methodology} \label{section_ME} 
In this section, we first present the overall pipeline of our MiniNet for unsupervised monocular depth estimation. Then, we elaborate the structure of the DepthNet, which can achieve a balance between high depth prediction accuracy and low model parameters via our proposed recurrent module and novel efficient upsample blocks. Finally, we introduce the associated loss functions for training our networks.

\begin{figure*}[h]
	\centering
	\makebox[\textwidth][c]{\includegraphics[scale=0.5]{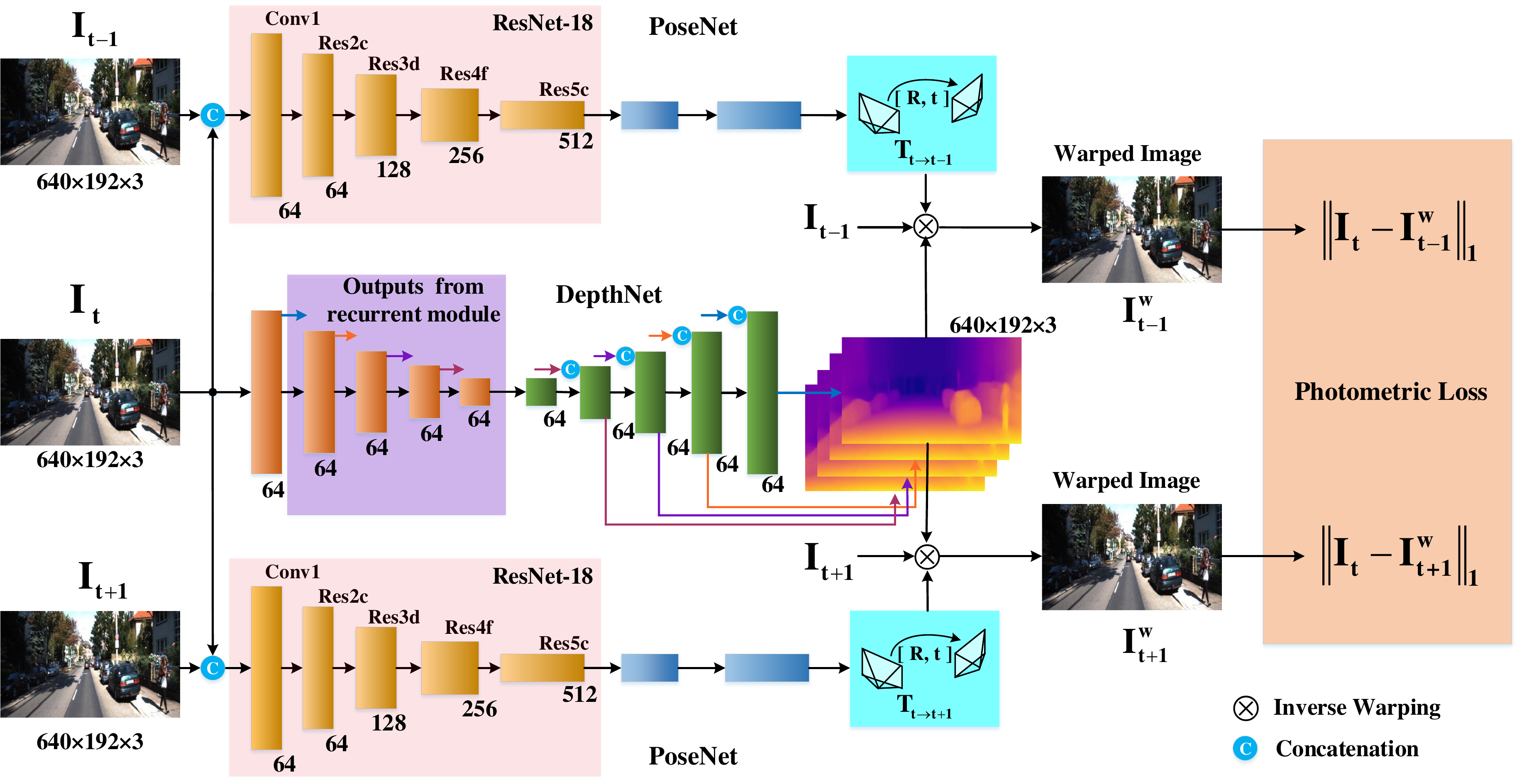}}
	\caption{Overview of our proposed MiniNet. The DepthNet only takes a single still image as input, while the two shared-weight PoseNets take the frame pair as input. The value below each feature map rectangle denotes the channel number, and the smaller height rectangle is the half size of the preceding one. Best viewed in color.}
	\label{fig:1}
\end{figure*}

\subsection{Overall pipeline of the MiniNet} \label{section_pipe} 
The proposed MiniNet is composed of a DepthNet and two shared-weight PoseNets as illustrated in Fig.~\ref{fig:1}. The DepthNet takes the target image to estimate the depth map, while the PoseNets take the adjacent two frames for camera ego-motion estimation. In the training stage, three consecutive video frames $ I_{t-1} $, $ I_{t} $, $ I_{t+1} $ are fed into the MiniNet, where the middle frame $ I_{t} $ is marked as the target image and the rest are source images. While in the inference stage, only the DepthNet is remained for single image depth prediction. The key idea in learning depth via an unsupervised manner is to enforce the geometry constraints of these consecutive video frames~\cite{RN291, RN435, RN575}. Our proposed MiniNet jointly produces the depth map $ D_{t} $ of target image $ I_{t} $ and the relative pose $ T_{t\rightarrow s} $ for each source image $ I_{s} $,  where $ s \in \left\{t-1, t+1\right\} $. Once given $ D_{t} $ and  $ T_{t\rightarrow s} $, the pixel $ p $ in the target image can be projected to the corresponding pixel $ p' $ in the source image by the following transformation: 
\begin{equation}
	p' = KT_{t\rightarrow s}D_{t}(p)K^{-1}p,
	\label{con:equation1}
\end{equation}
where $ K $ indicates the camera intrinsic matrix, which is a known parameter in this paper. As the value of $ p' $ is continuous, we adopt the differentiable bilinear interpolation strategy~\cite{RN591} to synthesize the target image $ I_{t} $ from source view $ I_{s} $, i.e.
\begin{equation}
	I_{s}^{\rm w} = I_{s}[p'],
	\label{con:equation2}
\end{equation}
where $ \left[ \cdot \right]  $ denotes the differentiable sampling operation. Under the assumption that the surface is Lambertian in the frame pair $ I_{t} $ and $ I_{s} $, i.e. the apparent brightness of corresponding pixels of two adjacent frames are remained uncharged, the photometric loss can be formulated as:
\begin{equation}
	\left\|I_{t} - I_{s}^{\rm w}\right\|_1,
	\label{con:equation3}
\end{equation}
where $ \left\| \cdot \right\|_1 $ denotes L1-norm.

Since our fundamental purpose is to obtain the real-time depth estimation and accurately estimated camera poses are important for accurate depth prediction, relatively deep architecture ResNet-18 is chosen as the encoder part of PoseNet, which is followed by four convolutional layers with ReLU nonlinear activation except for the last one. As shown in Fig.~\ref{fig:1}, the frame pair $ I_{s} $ and $ I_{t} $ is fed to PoseNet in the manner of concatenation along the color channels. Thus, the input channel of the first convolutional layer in ResNet-18 is modified to 6. To preserve the output value range, the initial weights of this convolution from pre-trained ResNet-18 on ImageNet~\cite{RN704} are halved. Each PoseNet can output 6-DoF transformation for each source image, which is scaled by 0.01 to facilitate regression learning as per~\cite{RN291, RN350}. The DepthNet of MiniNet has an extremely lightweight size, which is composed of an encoder with a recurrent module and a novel efficient decoder fusing the features from the corresponding encoder layer. We will elaborate on the structure of DepthNet in the following section.

\subsection{Structure of the lightweight DepthNet}
\begin{figure}[!t]
	\centering
	\includegraphics[scale=0.6]{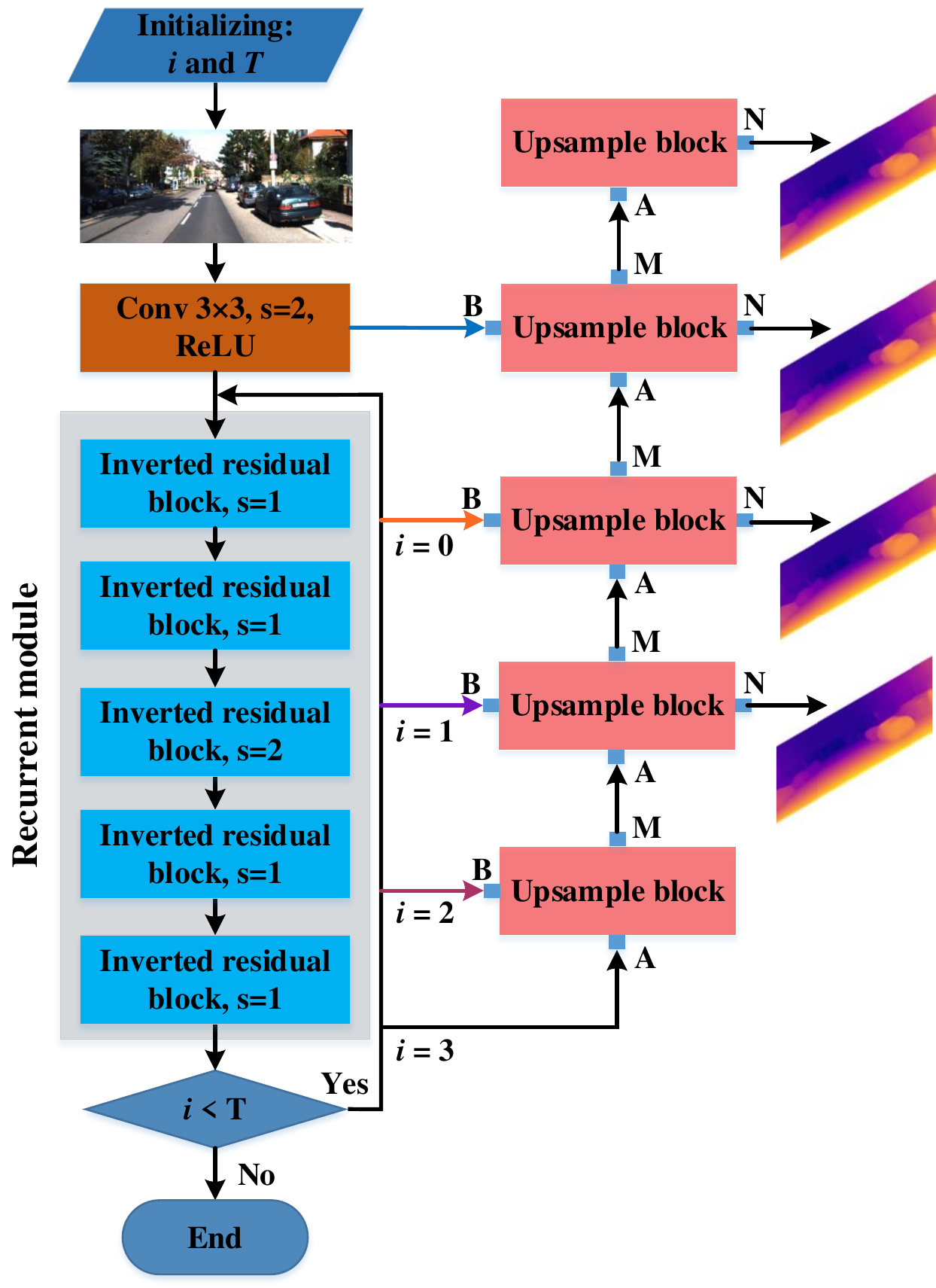}
	\caption{Schematic diagram of the DepthNet in our proposed MiniNet. The output feature maps from the first convolutional layer and the recurrent module will be skip-connected to the corresponding upsample blocks in the manner of concatenation. The DepthNet iteratively uses the recurrent module to generate multi-scale feature maps. $ i $ and $ T $ indicate the iteration time and total iteration number, respectively. s denotes the stride number of convolutional layer. The multi-scale disparity predictions will be bilinearly upsampled to the same spatial resolution of the input RGB image.}
	\label{fig:2}
\end{figure}

The structure of our proposed DepthNet is illustrated in Fig.~\ref{fig:2}, which consists of a recurrent module-based encoder and a novel efficient upsample block-based decoder.

\subsubsection{Recurrent module-based encoder}
As shown in Fig.~\ref{fig:2}, the encoder part of our DepthNet consists of a standard convolutional layer and a recurrent module. The same as the series of MobileNet (V1~\cite{RN697}, V2~\cite{RN650} and V3~\cite{RN698}), the first layer of our proposed encoder part is a standard $ 3 \times 3 $ convolution with the stride of 2 followed by ReLU activation. The output channel number $ c $ of the first layer is empirically set to 64. Our proposed recurrent module is composed of five inverted residual blocks, where the middle block has the stride of 2 and the rest have that of 1. Thus, the size of features will be halved via each iteration of the recurrent module. To achieve reusability, the input and output channel number of the recurrent module are designed to be identical, i.e. 64. Motivated by the series of ResNet~\cite{RN192}, the output stride of the encoder part is set to 32, i.e. the ratio of input image spatial resolution to the final output resolution of the encoder part. Thus, the halved feature maps from the first layer will iteratively pass through the recurrent module by four times, and the spatial feature size will be halved in each time. 

\begin{figure*}[!h]
	\centering
	\makebox[\textwidth][c]{\includegraphics[scale=0.5]{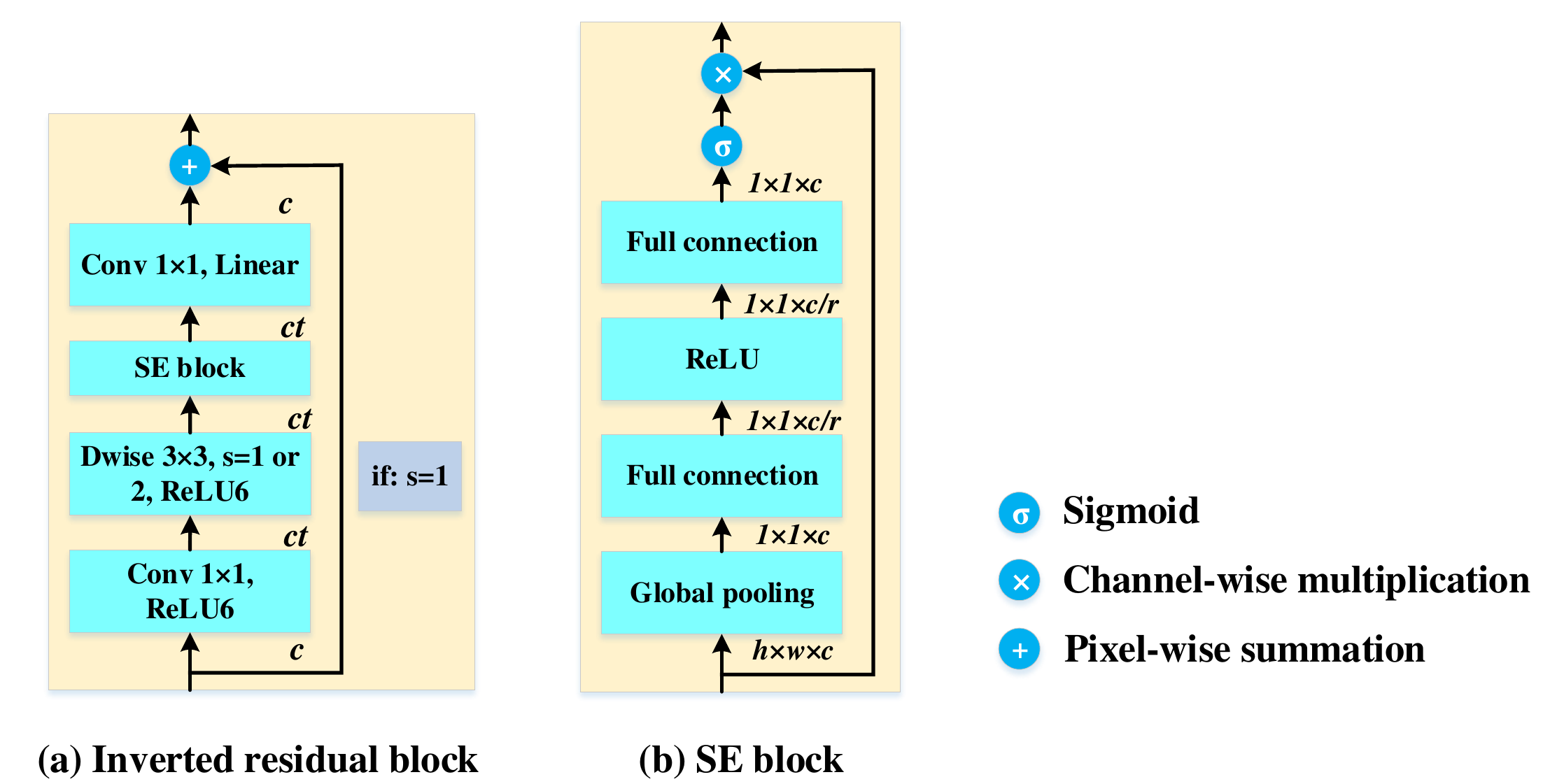}}
	\caption{Illustration of inverted residual block (a) and SE block (b). $ c $, $ t $ and s indicate the channel number of features, the expansion ratio and the stride of convolutional layer, respectively. Dwise indicates the depth-wise convolutional layer. $ r $ denotes the reduction ratio.}
	\label{fig:3-1}
\end{figure*}

The recurrent module is built upon the inverted residual block of MobileNetV3~\cite{RN698}, which has an inverted residual and linear bottleneck to alleviate the damage of feature maps caused by ReLU. The inverted residual block is composed of a $ 1 \times 1 $ (point-wise) convolution with ReLU6, a depth-wise (Dwise) convolution with ReLU6 and the stride of 1 or 2, a Squeeze-and-Excitation (SE) block~\cite{RN694} and a point-wise convolution without any nonlinear activation (linear bottleneck), as shown in Fig.~\ref{fig:3-1} (a). The channel number is expanded by the first point-wise convolution and then squeezed by the last one, which is inverted to the residual block of the original of ResNets~\cite{RN192}, where the channel number is first squeezed and then expanded. The shortcut connection is utilized between the input and output if the stride of the depth-wise convolution is equal to 1. The expansion ratio $ t $ of inverted residual block is set 2 or 4, which is the ratio of the output channel number to the input one of the first point-wise convolution. In order to strike a trade-off between model parameters and depth prediction accuracy, the expansion ratio of the first three inverted residual blocks is set to 2 and that of the last two blocks is set to 4 in our proposed recurrent module.

To boost the performance, SE block is adopted to regularize the feature maps of the recurrent module with a negligible increase of model parameters. SE block is composed of global pooling, two fully connected (FC) layers, ReLU non-linearity, sigmoid operation and channel-wise multiplication as shown in Fig.~\ref{fig:3-1} (b). As done in MobileNetV3, SE block is placed between the depth-wise convolution and the last point-wise convolution in the interior of the inverted residual block. According to~\cite{RN694}, the reduction ratio $ r $ of the full connection is set to 16 in our SE blocks. Our proposed MiniNet well combines the strengths of MobileNetV3 and recurrent neural networks (RNN), which enables the real-time inference and compact mode size for unsupervised monocular depth estimation.

\begin{figure*}[!h]
	\centering
	\makebox[\textwidth][c]{\includegraphics[scale=0.5]{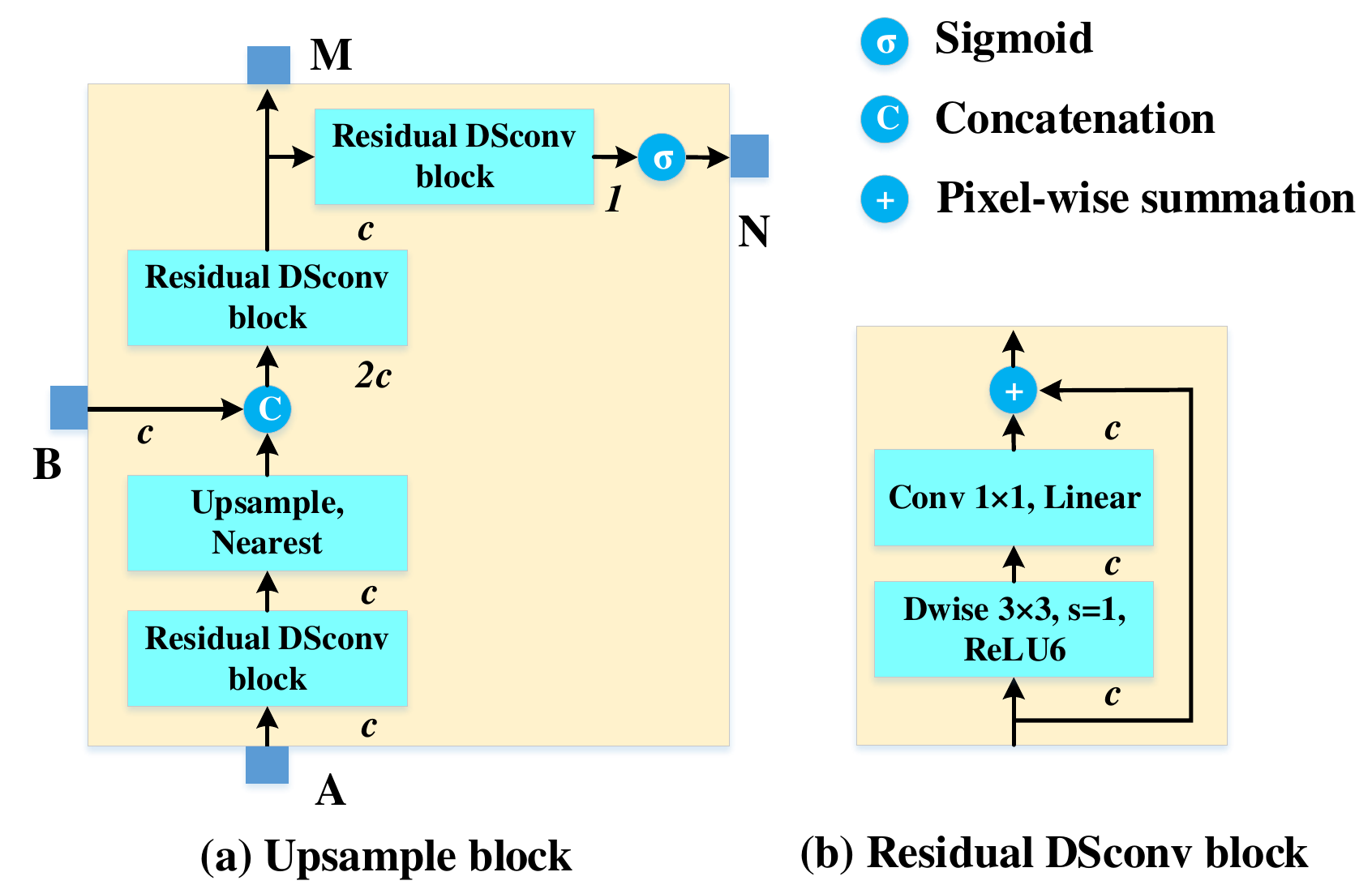}}
	\caption{Illustration of the proposed efficient upsample block (a) in the decoder part of DepthNet and residual DSconv block (b) in the upsample block. $ c $ and s indicate the channel number of features and the stride of convolutional layer, respectively. Dwise indicates the depth-wise convolutional layer.}
	\label{fig:3-2}
\end{figure*}

\subsubsection{Efficient upsample block-based decoder}
To meet the high-accuracy and real-time requirement, a novel efficient upsample block (as shown in Fig.~\ref{fig:3-2}) is designed to upsample and aggregate the output feature maps from the encoder. As shown in Fig.~\ref{fig:2}, the output feature maps from the first convolutional layer and the recurrent module will be skip-connected to the corresponding upsample blocks by concatenation. Unlike PyD-Net~\cite{RN572}, where a heavy decoder block with one deconvolution and four standard convolutions is used, our proposed lightweight upsample block is composed of three residual DSconv blocks, Nearest-upsample, concatenation, and sigmoid operations. These residual DSconv blocks are plug-in replacement of the standard convolutions, which consist of depth-wise and point-wise convolutions (i.e. depth-wise separable convolutions) with the shortcut connection between the input and output as shown in Fig.~\ref{fig:3-2} (b). All residual DSconv blocks have the input channel number of $ c $ except for the second one with $ 2c $ due to the concatenation operation. The third residual DSconv block followed by the sigmoid operation is exploited to attain multi-scale disparity maps $ P_{t} $, i.e. inverse depth maps. To improve the prediction accuracy, the multi-scale disparity maps are interpolated by bilinear-mode to the same spatial resolution of input RGB image. The multi-scale depth maps can be formulated as follows:
\begin{equation}
	D_{t} = 1/(aP_{t} + b),
	\label{con:equation4}
\end{equation}
where the constants $ a $ and $ b $ are set to 10 and 0.01 to constrain the predicted depth $ D_{t} $ to be always positive within a valid range.  

\subsection{Loss functions} 
In this section, the loss functions for training our proposed MiniNet are introduced. As explained in the previous section~\ref{section_pipe}, the fundamental loss function for unsupervised monocular depth estimation is the photometric reprojection loss. Motivated by~\cite{RN565}, the model-driven smoothness loss is appended to better explore the solution space of the disparity over the training stage instead of the standard smoothness loss. 

\noindent\textbf{Photometric loss.} We formulate a self-supervised signal from the image formation process via the photometric loss. Structured similarity (SSIM)~\cite{RN700} is a commonly-used metric for evaluating the quality of image predictions, which is adopted to measure the similarity between two image patches $ x $  and $ y $, and can be written as:
\begin{equation}
	{\rm SSIM}(x, y) = \dfrac{(2\mu_{x}\mu_{y} + c_{1})(2\sigma_{xy} + c_{2})}{(\mu_{x}^{2} + \mu_{y}^{2} + c_{1}) + (\sigma_{x}^{2} + \sigma_{y}^{2} + c_{2})},
	\label{con:equation5}
	\end{equation}
where $ \mu_{x}, \sigma_{x} $ are the local means and variances over the pixel neighborhood with $ c_{1}  = 0.01^{2} $ and $ c_{2} = 0.03^{2} $. Similar to~\cite{RN289, RN571} , the photometric loss is composed of L1-norm of the discrepancy between the synthesized and real images and SSIM, which is formulated as:
\begin{equation}
	\rho(I_{t}, I_{s}^{\rm w}) = \alpha\dfrac{1-{\rm SSIM}(I_{t}, I_{s}^{\rm w})}{2} + (1 - \alpha)\left\|I_{t} - I_{s}^{\rm w}\right\|_1, 
	\label{con:equation6}
\end{equation}
where the constant $ \alpha $ is commonly set to 0.85.

Since the target and source images are from different views, there are the occluded regions between the frame pair to some extent, which results in misleading information to the photometric loss. Fortunately, the occluded and dis-occluded regions result in the pixels from the target image not appearing in both the previous and next frames. Thus, the pixel-level minimum trick~\cite{RN435} is utilized to handle this problem instead of averaging the discrepancy errors from the source images. Our final photometric loss can be formulated as:
\begin{equation}
	\mathcal{L}_{ph} = \sum_{l} {\min_{s}}~\rho_{l}(I_{t}, I_{s}^{\rm w}) , 
	\label{con:equation7}
\end{equation}
where $ l $ denotes the multi-scale predictions.

\noindent\textbf{Model-driven smoothness loss.} To encourage the disparity outputs to be locally smooth meanwhile preserving sharp edge in the discontinuous regions, the edge-aware smoothness loss is usually adopted in self-supervised depth estimation, i.e.
\begin{equation}
	\mathcal{L}_{sm} =  \left| \partial_{x} d_{t}^{*} \right|e^{-\left\| \partial_{x} I_{t} \right\|_1} + \left| \partial_{y} d_{t}^{*} \right|e^{-\left\| \partial_{y} I_{t} \right\|_1},
	\label{con:equation8}
\end{equation}
where $ d_{t}^{*} = d_{t }/\overline{d_{t}} $ is the normalized disparity to remove the shrinking of predicted depth maps~\cite{RN350}. Furthermore, a spatial (pixel-level) and temporal (training-time) model-driven weight~\cite{RN565} will be adopted to better search the predicted depth space, and it can be written as:
\begin{equation}
	\beta_{i} = exp(-\dfrac{c\left\|I_{t}(i) - I_{s}^{\rm w}(i)\right\|_1}{\dfrac{1}{N}\sum_{i=1}^{N}\left\|I_{t}(i) - I_{s}^{\rm w}(i)\right\|_1}) , 
	\label{con:equation9}
\end{equation}
where $ N $ is the pixel number of the target image and $ c $ is empirically set to 10 for adjusting the range of $ \beta_{i} $ for the pixel $ i $. Thus, our model-driven smoothness loss can be formulated as:
\begin{equation}
	\mathcal{L}_{md} = \dfrac{1}{N}\sum_{i=1}^{N} \beta_{i}\mathcal{L}_{sm}^{i}.
	\label{con:equation10}
\end{equation}
The total loss function of MiniNet is composed of two terms, i.e.
\begin{equation}
	\mathcal{L} = \mathcal{L}_{ph} + \lambda \mathcal{L}_{md},
	\label{con:equation11}
\end{equation}
where the model-driven smoothness weight $ \lambda $ is empirically set to 0.001. Since the discrepancy is larger at the beginning of the training stage, the model-driven weight $ \beta_{i} $ is naturally small and the photometric loss is dominant to search solution space. With the proceeding of the training stage, the model-driven weight $ \beta_{i} $ increases, which will regularize the output disparities in a reasonable space.  

\section{Experiments} \label{section_EX}
To demonstrate the effectiveness of our approach, we evaluate it using the KITTI dataset~\cite{RN315}. In addition, we also use the Make3D dataset to show the generalization ability. In this section, we first introduce the used datasets. Then, we elaborate on the implementation details of our method. Finally, we present the experimental results of our approach.

\subsection{Datasets} 

KITTI dataset~\cite{RN315} is an outdoor dataset that contains 32 scenes for training and 29 scenes for testing using the Eigen split~\cite{RN268}. The RGB images and depth values are captured by car-mounted stereo cameras and rotating Velodyne laser scanner, respectively. Following the protocol of Godard et al.~\cite{RN435}, 39810 frames are used for training and 4424 frames are used for validation, where the static frames with mean optical flow magnitude less than 1 pixel are removed. 697 frames from 29 scenes are used for testing, and the Velodyne 3D points are reprojected into the left RGB camera using the given intrinsic and extrinsic parameters for evaluating depth estimation. The image resolution of each RGB frame and generated depth map is approximately $ 1226 \times 370 $ pixels. In this paper, the input RGB frames are resized to $ 640 \times 192 $ pixels for computational efficiency and maintaining the aspect ratio of the original RGB image. Besides, the KITTI odometry dataset~\cite{RN315} is used to evaluate the camera ego-motion accuracy of our proposed MiniNet, which contains 11 sequences (00-10) with ground-truth camera poses acquired by the IMU/GPS readings.


Make3D dataset~\cite{RN325} contains 400 training images and 134 testing images of outdoor scenes gathered by a custom 3D scanner.
Since we use it for evaluating the cross-dataset generalization ability of our proposed MiniNet, only the 134 testing images are used. The image resolution of input RGB images and ground-truth depth maps is $ 1704 \times 2272 $ pixels and $ 305 \times 55 $ pixels, respectively. Due to the different aspect ratio of Make3D with respect to KITTI, we use a central cropping of $ 2 \times 1 $ ratio proposed by Godard et al.~\cite{RN289} and thus attain a $ 1704 \times 852 $ crop centered on the image. Therefore, the height 55 of the ground-truth depth map is central-cropped to 21 proportionally. Following the previous works~\cite{RN291, RN289, RN565}, the errors are computed in the depth regions with ground-truth depth less than 70 meters.

\subsection{Implementation details} \label{section_DE}
Our proposed MiniNet is implemented in the publicly available PyTorch framework~\cite{RN711}. Batch normalization is only employed for the ResNet-18 modules of PoseNets in MiniNet. The weights of MiniNet are initialized by the method of Xavier~\cite{RN705} except for ResNet-18 modules of PoseNets, which are pre-trained on ImageNet. MiniNet is optimized by Adam~\cite{RN706} with the parameters $ \beta_{1} = 0.9 $ and $ \beta_{2} = 0.999 $ to improve the convergence rate. We train MiniNet on the RGB input resolution of $ 640 \times 192 $ pixels with a batch size of 6 for 40 epochs. The initial learning rate is set to $ 10^{-4} $, and reduced by 2 times every 30 epochs. Our proposed MiniNet is trained on a single Nvidia Geforce GTX 1080 Ti GPU with 11 GB memory, and it takes about 43 hours for training on the KITTI dataset. 

Data augmentation is performed online during the training stage to avoid over-fitting. We perform horizontal flips on three input frames with 50\% chance. Then, we perform color augmentation on brightness, contrast, saturation, and hue jitter with 50\% probability for these input frames. We uniformly sample from [0.8, 1.2] for brightness, contrast, saturation, and [0.9, 1.1] for hue jitter. After data augmentation, three input frames are divided by 255, and then normalized by the mean of 0.45 and standard deviation (std) of 0.225 according to ImageNet.

We quantitatively evaluate our MiniNet for unsupervised monocular depth estimation using several standard evaluation metrics as per the previous works~\cite{RN268, RN292, RN575}. Given $ N $ the total number of pixels of the target image and $ d_{t}^{i} $, $ \hat{d_{t}^{i}} $ the predicted depth and ground-truth depth values of pixel $ i $, we have:
 
(i) Mean absolute relative error, Abs Rel = $ \frac{1}{N} \sum_{i=1}^{N} \frac{\left| d_{t}^{i} - \hat{d_{t}^{i}}\right|}{\hat{d_{t}^{i}}} $;
 
(ii) Mean squared relative error, Sq Rel = $ \frac{1}{N} \sum_{i=1}^{N} \frac{\left| d_{t}^{i} - \hat{d_{t}^{i}}\right|^{2}}{\hat{d_{t}^{i}}} $;
 
(iii) Root mean squared error, RMSE = $ \sqrt{\frac{1}{N}\sum_{i=1}^{N} (d_{t}^{i} - \hat{d_{t}^{i}})^{2}} $;
 
(iv) Mean $ \log_{10} $ error, RMSE log = $ \sqrt{\frac{1}{N}\sum_{i=1}^{N} (\log_{10} d_{t}^{i} - \log_{10} \hat{d_{t}^{i}})^{2}} $;

(v) Accuracy within a threshold: the percentage of $ d_{t}^{i} $ s.t. $ \delta_{j} = \max\left(\frac{d_{t}^{i}}{\hat{d_{t}^{i}}}, \frac{\hat{d_{t}^{i}}}{d_{t}^{i}}\right) < 1.25^{j} $, where $ j $ = 1, 2, 3.

\begin{table}[!h]
	\centering
	\addtolength{\leftskip} {-2cm}
	\addtolength{\rightskip}{-2cm}
	\caption{Quantitative evaluation results on the KITTI dataset~\cite{RN315} using the Eigen split~\cite{RN268}. The referenced results are quoted from the corresponding papers respectively and are listed in a descending order of metric Abs Rel except for ours. '-' indicates that the result is not provided by the corresponding reference. For the method of Zhou et al.~\cite{RN682}, the parameters of HR-Net are evaluated.}
	{
		\resizebox{1.35\textwidth}{!}{
			\begin{tabular}{llccccccccccccc}
				\toprule[2pt]
				& \multicolumn{2}{c}{Setting} & & \multicolumn{4}{c}{Error (lower is better)} & & \multicolumn{3}{c}{Accuracy (higher is better)} & &\\
				\cline{2-3}	\cline{5-8}\cline{10-12} 
				Method & Cap     &Pose      &         & Abs Rel  &Sq Rel & RMSE  & RMSE log & &  $ \delta_{1} $   & $ \delta_{2} $ & $ \delta_{3} $  & & Parameters\\
				\midrule[1pt]
				Kuznietsov et al.~\cite{RN290}&0-80m      &\checkmark  & &0.308  &9.367  &8.700   &0.367    & & 0.752  &0.904  &0.952 & & 80.84 M\\
				Zhou et al.~\cite{RN291}      &0-80m      &\xmark      & &0.208  &1.768  &6.856   &0.283    & & 0.678  &0.885  &0.957 & & 34.20 M\\
				Mahjourian et al.~\cite{RN582}&0-80m      &\xmark      & &0.163  &1.240  &6.220   &0.250    & & 0.762  &0.916  &0.968 & & 31.59 M\\
				Yin et al.~\cite{RN293}       &0-80m      &\xmark      & &0.155  &1.296  &5.857   &0.233    & & 0.793  &0.931  &0.973 & & 58.45 M\\
				Poggi et al.~\cite{RN572}     &0-80m      &\checkmark  & &0.153  &1.363  &6.030   &0.252    & & 0.789  &0.918  &0.963 & & 1.972 M\\
				Pilzer et al.~\cite{RN297}    &0-80m      &\checkmark  & &0.152  &1.388  &6.016   &0.247    & & 0.789  &0.918  &0.965 & & 58.45 M\\
				Wang et al.~\cite{RN350}      &0-80m      &\xmark      & &0.151  &1.257  &5.583   &0.228    & & 0.810  &0.936  &0.974 & & 28.11 M\\
				Zou et al.~\cite{RN298}       &0-80m      &\xmark      & &0.150  &1.124  &5.507   &0.223    & & 0.806  &0.933  &0.973 & & 58.45 M\\
				Godard et al.~\cite{RN289}    &0-80m      &\checkmark  & &0.148  &1.344  &5.927   &0.247    & & 0.803  &0.922  &0.964 & & 31.60 M\\
				Ranjan et al.~\cite{RN575}    &0-80m      &\xmark      & &0.140  &1.070  &5.326   &0.217    & & 0.826  &0.941  &0.975 & & 80.88 M\\
				Elkerdawy et al.~\cite{RN645}      &0-80m      &\checkmark  & &0.136  &-      &5.891   &-        & & 0.827  &-      &-     & & 5.700 M\\
				Wong et al.~\cite{RN565}      &0-80m      &\checkmark  & &0.133  &1.126  &5.515   &0.231    & & 0.826  &0.934  &0.969 & & 20.81 M\\
				Gordon et al.~\cite{RN631}    &0-80m      &\xmark      & &0.128  &0.959  &5.230   &0.212    & & 0.845  &0.947  &0.976 & & 14.33 M\\
				Zhou et al.~\cite{RN682}      &0-80m      &\xmark      & &0.121  &0.837  &4.945   &0.197    & & 0.853  &0.955  &0.982 & & 34.16 M\\
				Tosi et al.~\cite{RN533}      &0-80m      &\checkmark  & &0.116  &0.986  &5.098   &0.214    & & 0.847  &0.939  &0.972 & & 42.50 M\\
				Godard et al.~\cite{RN435}    &0-80m      &\xmark      & &0.115  &0.903  &4.863   &0.193    & & 0.877  &0.959  &0.981 & & 14.84 M\\
				Casser et al.~\cite{RN668}    &0-80m      &\xmark      & &0.109  &0.825  &4.750   &0.187    & & 0.874  &0.958  &0.983 & & 14.33 M\\
				Ours                          &0-80m      &\xmark      & &0.141  &1.080  &5.264   &0.216    & & 0.825  &0.941  &0.976 & & 0.217 M\\		
				\midrule[1pt]
				Kuznietsov et al.~\cite{RN290}&1-50m      &\checkmark  & &0.262  &4.537  &6.182   &0.338    & & 0.768  &0.912  &0.955 & & 80.84 M\\
				Zhou et al.~\cite{RN291}      &0-50m      &\xmark      & &0.201  &1.391  &5.181   &0.264    & & 0.696  &0.900  &0.966 & & 34.20 M\\
				Garg et al.~\cite{RN288}      &1-50m      &\checkmark  & &0.169  &1.080  &5.104   &0.273    & & 0.740  &0.904  &0.962 & & 16.80 M\\
				Mahjourian et al.~\cite{RN582}&0-50m      &\xmark      & &0.155  &0.927  &4.549   &0.231    & & 0.781  &0.931  &0.975 & & 31.59 M\\
				Yin et al.~\cite{RN293}       &0-50m      &\xmark      & &0.147  &0.936  &4.348   &0.218    & & 0.810  &0.941  &0.977 & & 58.45 M\\
				Poggi et al.~\cite{RN572}     &0-50m      &\checkmark  & &0.145  &1.014  &4.608   &0.227    & & 0.813  &0.934  &0.972 & & 1.972 M\\
				Pilzer et al.~\cite{RN297}    &0-50m      &\checkmark  & &0.144  &1.007  &4.660   &0.240    & & 0.793  &0.923  &0.968 & & 58.45 M\\
				Godard et al.~\cite{RN289}    &0-50m      &\checkmark  & &0.140  &0.976  &4.471   &0.232    & & 0.818  &0.931  &0.969 & & 31.60 M\\
				Wong et al.~\cite{RN565}      &0-50m      &\checkmark  & &0.126  &0.832  &4.172   &0.217    & & 0.840  &0.941  &0.973 & & 20.81 M\\
				Ours                          &0-50m      &\xmark      & &0.135  &0.839  &4.067   &0.205    & & 0.838  &0.947  &0.978 & & 0.217 M\\ 
				
				\bottomrule[2pt]
		\end{tabular}}
	}
	\label{tab1}
\end{table} 

\subsection{Experimental results}
Firstly, we conduct quantitative and qualitative comparisons with other relevant works in this section. Secondly, we analyze the computational efficiency of our proposed MiniNet. Thirdly, we present the results of pose estimation on the KITTI odometry dataset. Fourthly, we conduct quantitative and qualitative experiments on the Make3D dataset to show the generalization ability of MiniNet. Finally, we perform ablation studies to demonstrate the effects of our proposed recurrent module and efficient upsample block. 

\subsubsection{Comparisons with other relevant works}
We present the evaluation results of MiniNet using the test split~\cite{RN268}. In Table~\ref{tab1}, we list the quantitative evaluation results of the relevant works trained on stereo image pairs or monocular video sequences. As per the relevant works~\cite{RN291, RN435, RN668}, the median scaling is adopted to align the estimations with the ground-truth depth in our MiniNet. The trainable parameters of the depth CNN of each work are listed in the rightmost column of Table~\ref{tab1}. As we can see from the upper part of Table~\ref{tab1}, our proposed MiniNet can attain the minimum of parameters (0.217 M), which is about 373 times smaller than that of the method of Ranjan et al.~\cite{RN575} (using DispResNet for depth estimation), and 0.871 megabytes (MB) model size with 32-bit floating point type. Compared with one of the first work of unsupervised depth estimation trained on monocular video~\cite{RN291}, our MiniNet obtains 32.2\% relative improvement on the metric Abs Rel and 158 times fewer parameters. Moreover, we compare our MiniNet to the method of Poggi et al.~\cite{RN572}, which is the most relevant work since it enables real-time depth estimation and has lightweight trainable parameters of 1.972 M. Although trained on monocular video sequences, our method outperforms the method of Poggi et al.~\cite{RN572} in all evaluation metrics with about 9 times smaller parameters. 

To compare with the first work of unsupervised depth estimation trained on stereo image pairs~\cite{RN288}, we also list the quantitative evaluation results with the cap of 50 meters (m) in the bottom part of Table~\ref{tab1}. Compared with Garg et al.~\cite{RN288}, our MiniNet achieves the best performance in all evaluation metrics with 20.1\% relative improvement on the metric Abs Rel and 77 times fewer parameters. Compared with Yin et al.~\cite{RN293}, where the ResFlowNet is utilized to improve the performance of depth estimation, our MiniNet also obtains the best performance in all evaluation metrics with 269 times fewer parameters. Besides, we compare our MiniNet with the recent method of Wong et al.~\cite{RN565}, whose model-driven adaptive weight is also adopted in our loss function. Our MiniNet achieves the better performance on the metrics of RMSE, RMSE log, $ \delta_{2} $, and $ \delta_{3} $ with 96 times fewer parameters.

\begin{figure*}[!h]
	\centering
	\makebox[\textwidth][c]{\includegraphics[scale=0.25]{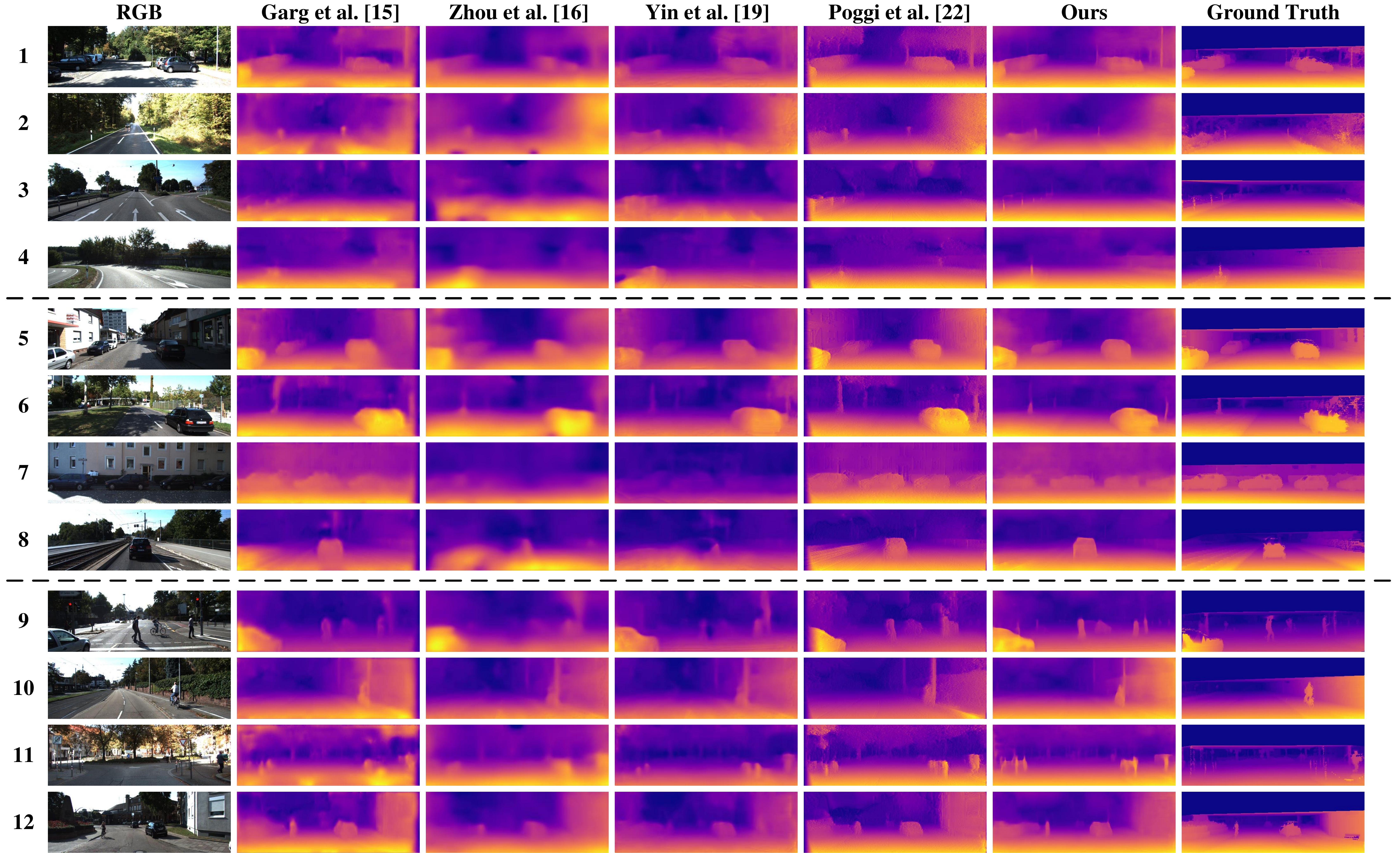}}
	\caption{Qualitative disparity results (i.e. the inverse depth maps) of different methods for twelve images in the test set of KITTI dataset using the plasma colormap. From the left to the right column: Input RGB image, Garg et al.~\cite{RN288}, Zhou et al.~\cite{RN291}, Yin et al.~\cite{RN293}, Poggi et al.~\cite{RN572}, Our results, and Ground-truth disparity map. The ground-truth disparity is interpolated from the sparse point cloud for better visualization.}
	\label{fig:4}
\end{figure*} 

To fairly compare the visual results of our MiniNet with the relevant works, we present the zoomed disparity maps for twelve images in Fig.~\ref{fig:4}. In the upper part of Fig.~\ref{fig:4}, i.e. the first four rows, our method can better capture thin structures, such as the lamppost in Row 1 and the traffic signs in Rows 2-4. In the middle part of Fig.~\ref{fig:4}, our method can delineate clearer object contours, such as cars in Rows 5-8. In the bottom part of Fig.~\ref{fig:4}, our method can accurately predict the position of the pedestrians in Rows 9, 11, and 12, and the cyclist in Row 10. It is significantly important for the applications of autopilot and UAV with security concerns. Although owning lightweight parameters about 0.217 M trained on monocular video sequences, our method provides decent visual results, which are comparable with the method of Poggi et al.~\cite{RN572} with 1.972 M parameters trained on stereo image pairs. Moreover, our visual results outperform that of other relevant works~\cite{RN288, RN291, RN293} with larger parameters exceeding 15 M.

\subsubsection{Analysis of computational efficiency}
In this section, we compare the computational burden of our MiniNet with that of Poggi et al.~\cite{RN572}. For a fair comparison, all the experiments are carried out on a desktop with an Intel E5-1630 CPU and a GTX 1080Ti GPU. The runtime is evaluated using a single GPU card or a single CPU card averaged over the test set of 697 forward passes. We present the computational performance at Full (F), Half (H), Quarter (Q) and Eighth (E) output sizes in Table~\ref{tab2}, as per Poggi et al.~\cite{RN572}. It should be noted that, in the upper part of Table~\ref{tab2}, the results of Poggi et al.~\cite{RN572} are retested for a fair comparison except for the metrics of RMSE and $ \delta_{1} $, which are directly quoted from the reference. These runtimes are close to the original data of Poggi et al.~\cite{RN572}.

\begin{table*}[h]
	\centering
	\addtolength{\leftskip} {-2cm}
	\addtolength{\rightskip}{-2cm}
	\caption{Computational efficiency study on the test set of the KITTI dataset. We conduct the experiments on the Full, Half, Quarter and Eighth output sizes. GPU and CPU indicate which the runtime tested on for a single forward pass. The best performances are highlighted in bold in each part. (The results of Poggi et al.~\cite{RN572} in Rows 1-3 are retested for a fair comparison except for the metrics of RMSE and $ \delta_{1} $, which are directly quoted from the reference.)}  
	{
		\resizebox{1.3\textwidth}{!}{                     
			\begin{tabular}{cllcccrccccccc}       
				\toprule[2pt]
			 & & & \multicolumn{2}{c}{Supervision}\\
				\cline{4-5}
		Row		&Model                         &Dataset    &Depth     &Pose    & Input Res.  & Output Res.  &Parameters  &Model size  &Flops      &GPU     &CPU  &RMSE      &$ \delta_{1} $\\
				\midrule[1pt]
		1		&Poggi et al.~\cite{RN572}     &KITTI         &\xmark      &\checkmark  & $ 512 \times 256 $    & $ 256 \times 128 $ (H)&1.972 M  &7.702 MB  &9.872 G   &11.15 ms    &0.116 s & \textbf{5.907}  &\textbf{0.801}\\
		2		&Poggi et al.~\cite{RN572}     &KITTI         &\xmark      &\checkmark  & $ 512 \times 256 $    & $ 128 \times 64 $ (Q) &1.874 M  &7.322 MB  &3.506 G   &9.867 ms    &0.045 s & 6.146  &0.787\\
		3		&Poggi et al.~\cite{RN572}     &KITTI         &\xmark      &\checkmark  & $ 512 \times 256 $    & $ 64 \times 32 $ (E)  &\textbf{1.763 M}  &\textbf{6.889 MB}  &\textbf{1.688 G}   &\textbf{9.289 ms}    &\textbf{0.028 s} & 7.222  &0.747\\
		\midrule[1pt]
		4       &Poggi et al.~\cite{RN572}     &KITTI         &\xmark      &\xmark      & $ 512 \times 256 $    & $ 256 \times 128 $ (H)&1.972 M  &7.702 MB   &9.872 G  &11.15 ms    &0.116 s  &5.759    &\textbf{0.833}\\
		5       &Poggi et al.~\cite{RN572}     &KITTI         &\xmark      &\xmark      & $ 512 \times 256 $    & $ 128 \times 64 $ (Q) &1.874 M  &7.322 MB   &3.506 G  &9.867 ms    &0.045 s  &5.882    &0.828\\
		6       &Poggi et al.~\cite{RN572}     &KITTI         &\xmark      &\xmark      & $ 512 \times 256 $    & $ 64\times 32 $ (E)   &1.763 M  &6.889 MB   &\textbf{1.688 G}  &\textbf{9.289 ms}    &\textbf{0.028 s}  &6.205    &0.812\\
		7		&Ours        &KITTI       &\xmark    &\xmark   & $ 512 \times 256 $    & $ 512 \times 256 $ (F)&0.217 M  &0.871 MB  &8.235 G  &19.72 ms    &0.216 s &\textbf{5.182}   &0.827\\
		8       &Ours        &KITTI       &\xmark    &\xmark   & $ 512 \times 256 $    & $ 256 \times 128 $ (H)&0.208 M  &0.821 MB  &6.629 G  &16.37 ms    &0.156 s &5.187            &0.827\\
		9       &Ours        &KITTI       &\xmark    &\xmark   & $ 512 \times 256 $    & $ 128 \times 64 $ (Q) &0.193 M  &0.774 MB  &5.918 G  &15.01 ms    &0.134 s &5.213            &0.824\\
		10      &Ours        &KITTI       &\xmark    &\xmark   & $ 512 \times 256 $    & $ 64 \times 32 $ (E)  &\textbf{0.179 M}   &\textbf{0.717 MB}  &5.740 G  &14.67 ms  &0.125 s &5.302   &0.819\\
		\midrule[1pt]
		11		&Ours                          &KITTI         &\xmark      &\xmark      & $ 640 \times 192 $    & $ 640 \times 192 $ (F)&0.217 M  &0.871 MB  &7.720 G   &18.57 ms    &0.205 s & 5.264  &\textbf{0.825}\\
		12		&Ours                          &KITTI         &\xmark      &\xmark      & $ 640 \times 192 $    & $ 320 \times 96 $ (H) &0.208 M  &0.821 MB  &6.215 G   &15.58 ms    &0.143 s & \textbf{5.252}  &0.823\\
		13		&Ours                          &KITTI         &\xmark      &\xmark      & $ 640 \times 192 $    & $ 160 \times 48 $ (Q) &0.193 M  &0.774 MB  &5.548 G   &14.40 ms    &0.120 s & 5.262  &0.821\\
		14		&Ours                          &KITTI         &\xmark      &\xmark      & $ 640 \times 192 $    & $ 80 \times 24 $ (E)  &\textbf{0.179 M}  &\textbf{0.717 MB}  &\textbf{5.381 G}   &\textbf{14.00 ms}    &\textbf{0.113 s} & 5.337  &0.814\\
				\midrule[1pt]
		15		&Ours (medium)                 &KITTI         &\xmark      &\xmark      & $ 640 \times 192 $    & $ 640 \times 192 $ (F)&0.110 M  &0.449 MB  &3.729 G   &14.10 ms    &0.126 s & \textbf{5.455}  &\textbf{0.799}\\
		16		&Ours (medium)                 &KITTI         &\xmark      &\xmark      & $ 640 \times 192 $    & $ 80 \times 24 $ (E)  &0.072 M  &0.295 MB  &1.391 G   &9.898 ms    &0.036 s & 5.540  &0.790\\
		17		&Ours (small)                  &KITTI         &\xmark      &\xmark      & $ 640 \times 192 $    & $ 640 \times 192 $ (F)&0.091 M  &0.371 MB  &3.366 G   &13.59 ms    &0.119 s & 5.581  &0.792\\
		18		&Ours (small)                  &KITTI         &\xmark      &\xmark      & $ 640 \times 192 $    & $ 80 \times 24 $ (E)  &\textbf{0.053 M}  &\textbf{0.217 MB}  &\textbf{1.028 G}   &\textbf{9.136 ms}    &\textbf{0.027 s} & 5.645  &0.781\\ 
				
				\bottomrule[2pt]
		\end{tabular}}
	}
	\label{tab2}
\end{table*} 

To better compare with the method of Poggi et al.~\cite{RN572}, we train both our model and theirs on monocular video sequences with the input size of $ 512 \times 256 $ as shown in the second part of Table~\ref{tab2}. Our method attains better performance on RMSE at all the output resolutions and $ \delta_{1} $ at eighth resolution with respect to the method of Poggi et al.~\cite{RN572}. As we can see from Rows 7 and 10, our methods obtain real-time inference with about 51 fps at full resolution ($ 512 \times 256 $ pixels) and 68 fps at the eighth resolution on a GTX 1080Ti GPU card. At half resolution (Rows 4 and 8), the flops of our method is about 67.2\% of that of Poggi et al.~\cite{RN572} and at the same time has about 9 times fewer parameters and model size. Although the flops of our MiniNet (Row 8) is smaller than that of the associated one in Row 4, our runtimes for a single forward pass are larger than that of Poggi et al.~\cite{RN572} on both GPU and CPU cards. We conjecture that depth-wise convolutions extensively used in our models are not fully optimized in commonly-used deep learning framework we used. At quarter resolution (Rows 5 and 9), despite of higher flops, our method still outperforms Poggi et al.~\cite{RN572} on the metrics of model volume and RMSE. At eighth resolution (Rows 6 and 10), the flops of Poggi et al.~\cite{RN572} is about 3 times fewer than that of ours, but our method attains 14.6 \% and 0.9 \% relative improvements on the metrics of RMSE and $ \delta_{1} $ with about 10 times fewer model parameters.

Unlike the method of Poggi et al.~\cite{RN572} with $ 512 \times 256 $ input size, the input size of $ 640 \times 192 $ is mainly adopted in our models to preserve the aspect ratio of the original RGB image with about $ 1226 \times 370 $ pixels. The corresponding results are shown in the third part of Table~\ref{tab2} (i.e. Rows 11-14). Our method achieves real-time inference about 54 fps at full resolution output size ($ 640 \times 192 $ pixels) and 71 fps at the eighth resolution on a GTX 1080Ti GPU card, as shown in Rows 11 and 14. Although the runtimes of our model trained on the input size of $ 640 \times 192 $ are faster than that of our model trained on the input size of $ 512 \times 256 $, the latter obtains better performance on depth prediction accuracy as shown in Rows 7-14.  It is because that the image with $ 512 \times 256 $ has more pixels and thus can capture more details of the scenes.

\begin{figure*}[!h]
	\centering
	\makebox[\textwidth][c]{\includegraphics[scale=0.28]{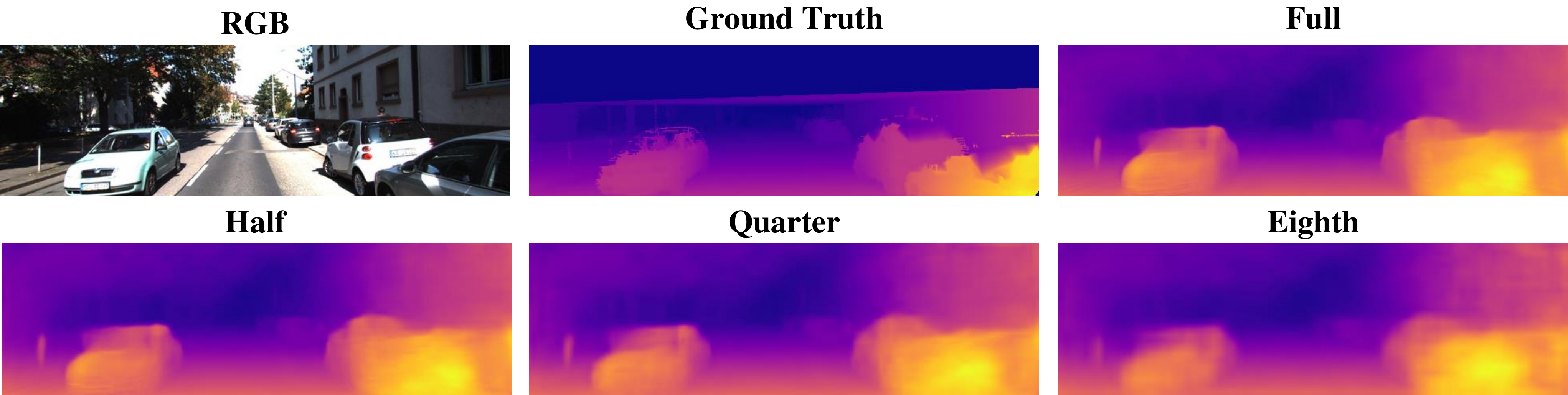}}
	\caption{Qualitative disparity comparisons on a typical test KITTI image with the four output sizes (Full, Half, Quarter, and Eighth).}
	\label{fig:5-1}
\end{figure*}

\begin{figure}[!h]
	\centering
	\includegraphics[scale=0.6]{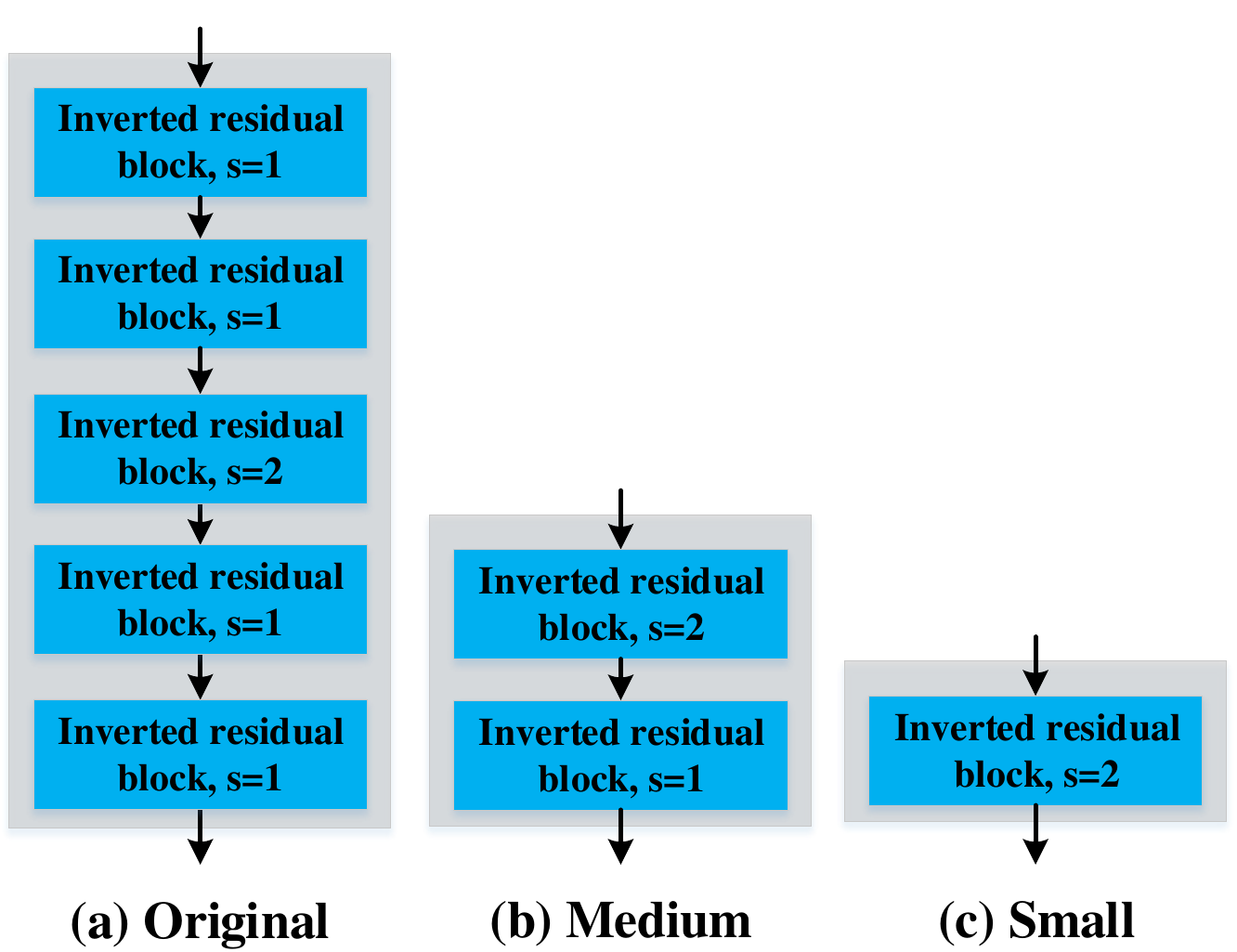}
	\caption{Three versions of the recurrent module in our DepthNet. (a) Original version from Fig.~\ref{fig:2}. (b) Medium version with two inverted residual blocks, where the stride of the first one is set to 2. (c) Small version with only one inverted residual block using the stride of 2.}
	\label{fig:6}
\end{figure}

\begin{figure*}[t]
	\centering
	\makebox[\textwidth][c]{\includegraphics[scale=0.28]{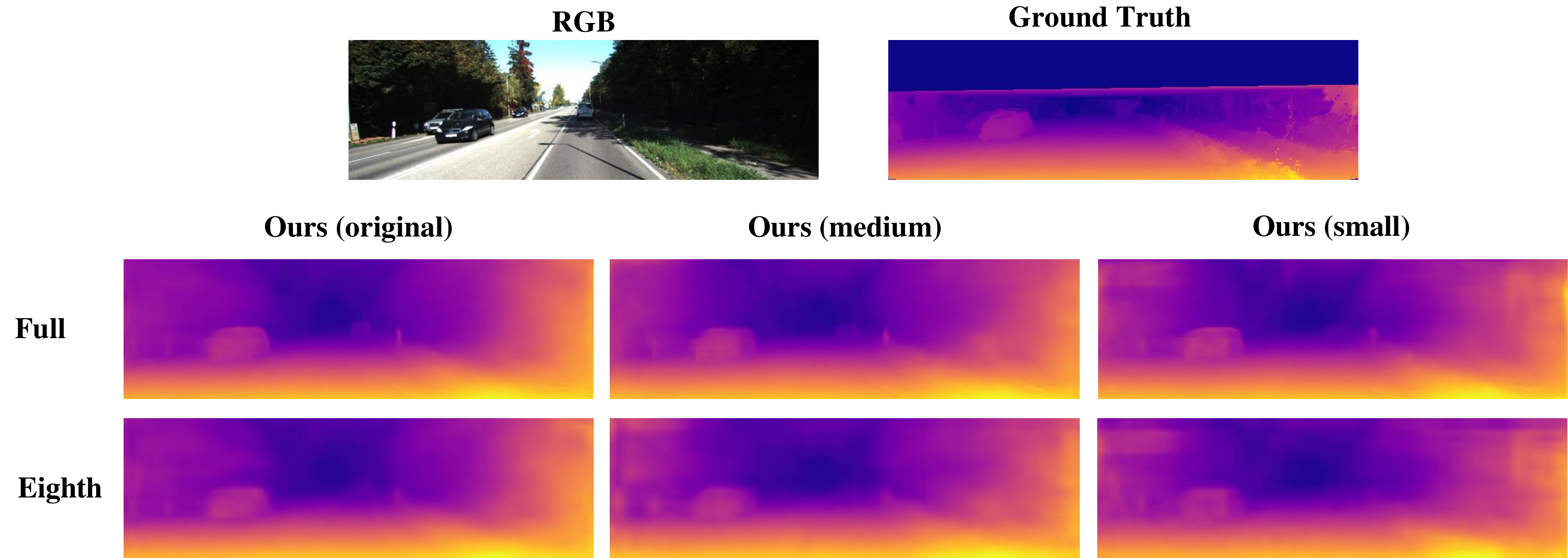}}
	\caption{Qualitative disparity comparisons of our proposed recurrent modules of Fig.~\ref{fig:6}, i.e. original, medium and small versions for full and eighth output sizes.}
	\label{fig:5-2}
\end{figure*}
  
Owing to the same size multi-scale self-supervised signals, our method is capable of obtaining the similar accuracy on different output resolutions. The visual results of multi-scale disparity outputs are shown in Fig.~\ref{fig:5-1}. Except for the eighth output, we can observe almost the same visual effects on other resolutions, especially for the traffic sign and car in the left of the RGB image. To obtain the smaller flops with small accuracy degrade, the eighth resolution is chosen as the final output size. Unlike the models of Poggi et al.~\cite{RN572}, the computation burden of our models mainly comes from the encoder, instead of the decoder. To realize the real-time performance on a single CPU card, the medium and small versions of recurrent module are proposed to further reduce the flops as shown in Fig.~\ref{fig:6}. For the medium version, two inverted residual blocks are exploited as shown in Fig.~\ref{fig:6} (b), where the stride of the first one is set to 2 for obtaining lower flops by reducing the feature sizes in advance and the expansion ratio of them are both set to 2. For the small version, only one inverted residual block is utilized with the stride and expansion ratio of 2 as illustrated in Fig.~\ref{fig:6} (c). The computational analysis of these two new versions is described in the bottom part of Table~\ref{tab2}. For both medium and small versions, the eighth output can significantly reduce the requirement of flops by about three times with respect to the full output. Furthermore, there are a small decrease on the metrics of RMSE and $ \delta_{1} $. At eighth resolution, medium and small versions respectively obtain 59.8\% and 70.4\% reduction ratio of parameters with respect to the original version in Fig.~\ref{fig:6} (a). Besides, our small version attains 0.053 M parameters and 1.028 G flops, which are about 33 times and 1.6 times fewer than the corresponding one (Row 3 in Table~\ref{tab2}) of Poggi et al~\cite{RN572} respectively. Finally, our small version achieves real-time performances on both GPU and CPU cards, which are about 110 fps and 37 fps respectively. The corresponding visual results of our medium and small versions are illustrated in Fig.~\ref{fig:5-2}. 

\begin{table}[t]
	\centering
	\caption{Absolute trajectory error (ATE) on the test set of the KITTI odometry dataset averaged over 5-frame snippets with standard deviation in meters (lower is better).}
	{
		\resizebox{0.5\textwidth}{!}{                     
			\begin{tabular}{lcc}       
				\toprule[2pt]
				Method                          &Seq. 09                &Seq. 10       \\
				\midrule[1pt]
				ORB-SLAM (short)~\cite{RN499}   &$ 0.064 \pm 0.141 $    &$ 0.064 \pm 0.130 $ \\
				Wang et al.~\cite{RN350}        &$ 0.045 \pm 0.108 $    &$ 0.033 \pm 0.074 $ \\
				Zhou et al.~\cite{RN291}        &$ 0.021 \pm 0.017 $    &$ 0.020 \pm 0.015 $ \\
				Godard et al.~\cite{RN435}      &$ 0.017 \pm 0.008 $    &$ 0.015 \pm 0.010 $ \\
				Zou et al.~\cite{RN298}         &$ 0.017 \pm 0.007 $    &$ 0.015 \pm 0.009 $ \\
				Zhou et al.~\cite{RN682}        &$ 0.015 \pm 0.007 $    &$ 0.015 \pm 0.009 $ \\
				ORB-SLAM (full)~\cite{RN499}    &$ 0.014 \pm 0.008 $    &$ 0.012 \pm 0.011 $ \\ 
				Mahjourian et al~\cite{RN582}   &$ 0.013 \pm 0.010 $    &$ 0.012 \pm 0.011 $ \\
				Yin et al.~\cite{RN293}         &$ 0.012 \pm 0.007 $    &$ 0.012 \pm 0.009 $ \\
				Ranjan et al.~\cite{RN575}      &$ 0.012 \pm 0.007 $    &$ 0.012 \pm 0.008 $ \\
				Wang et al.~\cite{RN570}        &$ 0.012 \pm 0.006 $    &$ 0.013 \pm 0.008 $ \\
				Bozorgtabar et al.~\cite{RN680} &$ 0.011 \pm 0.007 $    &$ 0.011 \pm 0.015 $ \\
				Casser et al.~\cite{RN668}      &$ 0.011 \pm 0.006 $    &$ 0.011 \pm 0.010 $ \\
				Chen et al.~\cite{RN630}        &$ 0.011 \pm 0.006 $    &$ 0.011 \pm 0.009 $ \\
				Gordon et al.~\cite{RN631}      &$ 0.010 \pm 0.016 $    &$ 0.007 \pm 0.009 $ \\
				Ours                            &$ 0.020 \pm 0.010 $    &$ 0.017 \pm 0.010 $ \\			
				\bottomrule[2pt]
		\end{tabular}}
	}
	\label{tab3}
\end{table}

\subsubsection{Evaluation of pose estimation}
For completeness, we provide the performance of MiniNet on pose estimation (i.e. camera ego-motion) following the official KITTI odometry split~\cite{RN291}, since the DepthNet and PoseNets are learned jointly and their accuracy are interrelated. We first train our MiniNet on the sequences 00-08, and then test it on sequences 09-10. The total testing sequence lengths are 1702 and 918 meters, respectively. Here, we compare the pose estimation of MiniNet with the traditional popular SLAM system ORB-SLAM~\cite{RN499}. We present two variants: ORB-SLAM (full), which takes the whole sequence as input allowing loop closure detection and re-localization, and ORB-SLAM (short), which only takes 5-frame snippets as input. The evaluation metric of odometry is absolute trajectory error (ATE) averaged over 5-frame snippets. Since the input of our PoseNet is two frames, we accumulate the estimations of four-pairs from each set of 5-frame snippets to obtain local trajectories. As per Zhou et al.~\cite{RN291}, we align the estimated local trajectory with the associated ground-truth to address the scale ambiguity during evaluation. The pose estimation results are summarized in Table~\ref{tab3} with a descending order of ATE except for ours. As we can see in Table~\ref{tab3}, our PoseNet shows competitive performance with ORB-SLAM and other unsupervised learning methods, especially the method of Godard et al.~\cite{RN435}, where two frames were also fed to predict camera ego-motion. The results demonstrate that our lightweight DepthNet is able to favorably provide support for camera ego-motion estimation.

\begin{table}[h]
	\centering
	\caption{Quantitative evaluation results on the Make3D dataset~\cite{RN325}. To demonstrate generalization ability of our MiniNet, we do not use any of the Make3D data for training, and directly adopt the model trained on the KITTI dataset to the test set of Make3D. Following the evaluation protocol of Godard et al.~\cite{RN289}, the errors are only computed for the pixels in a central image crop of $ 2 \times 1 $ ratio with ground-truth depth less than 70 meters.}
	{
		\resizebox{0.7\textwidth}{!}{                     
			\begin{tabular}{lccccc}       
				\toprule[2pt]
				Method                       & Supervision    &Abs Rel  &Sq Rel      &RMSE     &RMSE log\\
				\midrule[1pt]
				Train set mean               & depth          &0.876    &13.98       &12.27    &0.307   \\
				Karsch et al.~\cite{RN322}   & depth          &0.428    &5.079       &8.389    &0.149   \\
				Liu et al.~\cite{RN323}      & depth          &0.475    &6.562       &10.05    &0.165   \\
				Laina et al.~\cite{RN284}    & depth          &0.204    &1.840       &5.683    &0.084   \\
				\midrule[1pt]
				Godard et al.~\cite{RN289}   & pose           &0.544    &10.94       &11.76    &0.193   \\
				Wong et al.~\cite{RN565}     & pose           &0.427    &8.183       &11.78    &0.156   \\
				Wang et al.~\cite{RN350}     & none           &0.387    &4.720       &8.090    &0.204   \\
				Zhou et al.~\cite{RN291}     & none           &0.383    &5.321       &10.47    &0.478   \\
				Zou et al.~\cite{RN298}      & none           &0.331    &2.698       &6.890    &0.416   \\
				Bozorgtabar~\cite{RN680}     & none           &0.330    &2.692       &6.850    &0.412   \\
				Godard et al.~\cite{RN435}   & none           &0.322    &3.589       &7.417    &0.163   \\
				Zhou et al.~\cite{RN682}     & none           &0.318    &2.288       &6.669    &-       \\
				Ours                         & none           &0.398    &5.167       &8.534    &0.192   \\			
				\bottomrule[2pt]
		\end{tabular}}
	}
	\label{tab4}
\end{table}

\subsubsection{Generalization test on Make3D}
\begin{figure}[!h]
	\centering
	\includegraphics[scale=0.32]{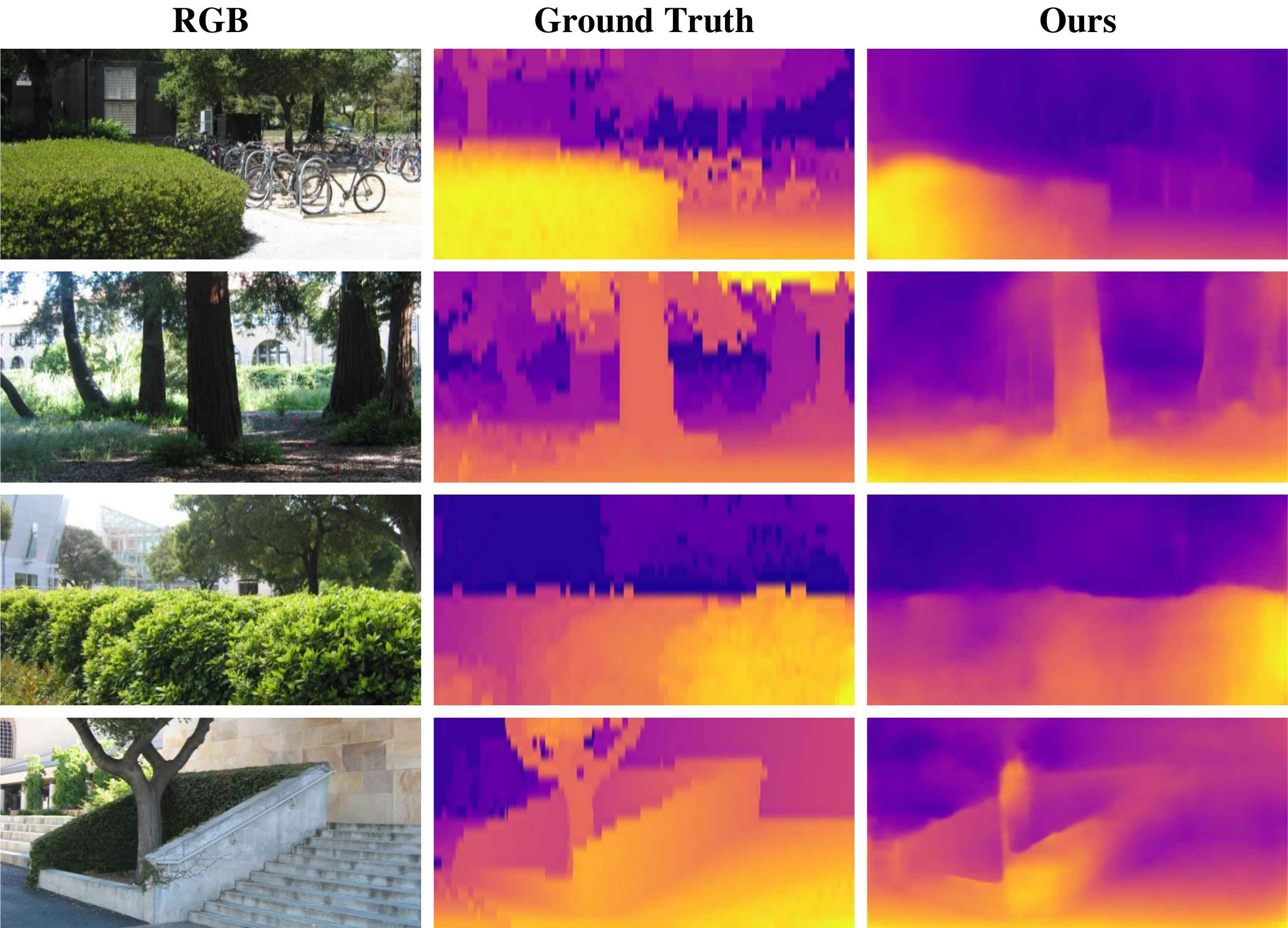}
	\caption{Qualitative disparity results on the Make3D test set. Note that our model is only trained on the KITTI dataset without any fine-tuning for Make3D.}
	\label{fig:7}
\end{figure}

To illustrate the generalization ability of our MiniNet on general scenes, we directly apply our model trained on the KITTI dataset to the Make3D dataset without any fine-tuning on it. The results of the supervised methods trained on the Make3D dataset with ground-truth depth are listed at the upper part of Table~\ref{tab4}, whereas the results of the unsupervised methods trained on the KITTI dataset are listed at the bottom part of Table~\ref{tab4} for better comparison. As we can see from Table~\ref{tab4}, our MiniNet, even without pre-training on the Cityscapes dataset~\cite{RN616}, achieves comparable results with respect to the supervised and unsupervised methods, which have much more parameters than ours. As shown in Fig.~\ref{fig:7}, our method, with extremely small number of parameters, is able to reasonably capture scene geometry structure such as tree trunks and shrubs.

\subsubsection{Ablation study}
To better demonstrate how the recurrent module and efficient lightweight upsample block contribute to the overall performance in unsupervised monocular depth estimation, we evaluate two variants of our method for the ablation study and the results are presented in Table~\ref{tab5}. In this table, ``w/o reuse" refers that four original modules as shown in Fig.~\ref{fig:6} (a) are stacked to fulfill the function instead of reusing the recurrent module four times, and ``w/o lightweight decoder" indicates that the standard convolutions are used to replace residual DSconv blocks in Fig.~\ref{fig:3-2} (a). As we can see from Table~\ref{tab5}, our recurrent module can significantly reduce the parameters and model size of our DepthNet. Compared with the ``w/o reuse" method, our MiniNet shows competitive performance on the unsupervised monocular depth estimation with approximately three times fewer parameters and model size. Meanwhile, our MiniNet can achieve nearly identical runtime and reduce more than 10\% memory usage of the ``w/o reuse" method on the Raspberry Pi 3 (ARM v8 processor Cortex-A53 1.2GHz) with 1 gigabyte (GB) memory. Thus, our recurrent module can help embedded devices save storage and memory spaces for executing other tasks. As for our proposed lightweight upsample block, it can effectively alleviate the storage requirements and reduce the runtime, which helps our DepthNet attain about tripled improvement on the parameters, model size, and flops. Meanwhile, it runs 1.7 times faster on the Raspberry Pi 3 with nearly identical memory usage and very limited performance degradation on depth prediction accuracy. By adopting the proposed recurrent module and the novel efficient upsample block, the parameters of our MiniNet are approximately 9 times fewer than that of Poggi et al.~\cite{RN572}. When the eighth output resolution and the small version of MiniNet are further adopted, our model can achieve about 2 fps with 148 MB memory usage and 0.053 M parameters on the Raspberry Pi 3.

\begin{table*}[h]
	\caption{Quantitative evaluation results of different variants of our approach on the KITTI dataset for the ablation study with the cap of 80 meters and the full output resolution of $ 640 \times 192 $. The memory usage and runtime are reported by testing on the Raspberry Pi 3.}
	{
		\resizebox{1\textwidth}{!}{
			\begin{tabular}{llllclcccccccc}       
				\toprule[2pt]
				&   &  & & & \multicolumn{4}{c}{Error (lower is better)} & & \multicolumn{3}{c}{Accuracy (higher is better)}\\
				\cline{7-10}\cline{12-14}
				Method                          &Parameters &Model size &Flops    &Memory usage &Runtime      &Abs Rel  &Sq Rel    &RMSE   &RMSE log  & &$ \delta_{1} $  &$ \delta_{2} $  & $ \delta_{3} $\\
				\midrule[1pt]
				w/o reuse                       &0.656 M     &2.673 MB   &7.720 G  &178 MB      &5.351 s   &0.131    &1.004     &5.070  &0.208     & &0.845           &0.945           &0.976 \\
				w/o lightweight decoder         &0.667 M     &2.685 MB   &23.45 G  &157 MB      &9.078 s   &0.134    &1.012     &5.110  &0.210     & &0.840           &0.947           &0.977 \\
				Our MiniNet                     &0.217 M     &0.871 MB   &7.720 G  &159 MB      &5.321 s   &0.141    &1.080     &5.264  &0.216     & &0.825           &0.941           &0.976 \\
				\bottomrule[2pt]
		\end{tabular}}
	}
	\label{tab5}
\end{table*}

\section{Conclusions} \label{section_CO}
To the best of our knowledge, we are the first to introduce a lightweight real-time deep neural network (named MiniNet) trained on monocular video for unsupervised depth estimation. The core of the proposed MiniNet is the DepthNet. A recurrent module is therein proposed to reduce the parameters of the DepthNet, which enables the encoder of the DepthNet to achieve both high accuracy and low parameters. Besides, a novel efficient upsample block is proposed for pixel-level depth estimation, where the depth-wise separable convolution with the shortcut is exploited. We conduct extensive experiments on the KITTI dataset to demonstrate the effectiveness and efficiency of our proposed MiniNet. Our approach with minimal parameters achieves competitive depth prediction accuracy to the methods trained on stereo image pairs or monocular video sequences. Moreover, our small version attains the real-time performance with about 110 fps on a single GPU card and 37 fps on a single CPU card, as well as about 2 fps on a Raspberry Pi 3. Meanwhile, it has both higher depth prediction accuracy and fewer parameters than that of the state-of-the-art lightweight unsupervised depth estimation methods.  Due to the lightweight and real-time merits, our proposed method can facilitate many applications in the photogrammetry and remote sensing communities, such as visual odometry, image localization, and height estimation, which are worth further investigation in the future studies.

\section*{Acknowledgements}
This work was partially supported by the Education Department of Guangdong Province, PR China, under project No 2019KZDZX1028.

\section*{References}

\bibliography{mybibfile}

\end{document}